\pdfoutput=1
\documentclass[11pt]{article}

\usepackage[final]{acl}

\usepackage{times}
\usepackage{latexsym}

\usepackage[T1]{fontenc}
\usepackage[utf8]{inputenc}

\usepackage{microtype}

\usepackage{inconsolata}

\usepackage{graphicx}
\usepackage{booktabs}

\usepackage{enumitem}  %
\usepackage{xcolor}    %
\usepackage{tcolorbox} %
\usepackage{soul}   
\newcommand{\hlc}[2][yellow]{{%
    \colorlet{foo}{#1}%
    \sethlcolor{foo}\hl{#2}}%
}

\usepackage{multirow}

\usepackage[]{todonotes}

\definecolor{ocr}{HTML}{00C8FF}
\definecolor{ocr}{HTML}{009900}
\definecolor{owenColor}{rgb}{0.0, 1.0, 0.8}
\definecolor{owenurgentColor}{rgb}{0.0, 1.0, 0.8}
\definecolor{jackColor}{rgb}{1.0, 0.6, 0.4}
\definecolor{susanColor}{rgb}{0.5, 0.2, 0.8}
\definecolor{weilingColor}{rgb}{0.2, 0.6, 0.2}
\definecolor{panosColor}{rgb}{0.8, 0.2, 0.3}
\definecolor{gregColor}{rgb}{0.2, 0.2, 0.8}
\definecolor{ForestGreen}{RGB}{20,150,20}

\title{LVLMs are Bad at Overhearing Human Referential Communication}
\author{
Zhengxiang Wang\textsuperscript{1,2} \hspace{.1cm} 
Weiling Li\textsuperscript{3} \hspace{.1cm} 
Panagiotis Kaliosis\textsuperscript{4} \hspace{.1cm} 
Owen Rambow\textsuperscript{1,2} 
\hspace{.1cm} 
Susan E. Brennan\textsuperscript{3} \\
\textsuperscript{1} Department of Linguistics \hspace{.1cm} 
\textsuperscript{2} Institute for Advanced Computational Science \\
\textsuperscript{3} Department of Psychology \hspace{.1cm} 
\textsuperscript{4} Department of Computer Science, Stony Brook University \\
\texttt{zhengxiang.wang@stonybrook.edu}
}

\begin{document}
\maketitle
\begin{abstract}
%Understanding the referring expressions people generate during spontaneous conversational interaction is an important ability for an embodied agent to carry out tasks in the real world. Doing this requires integrating and  understanding language, vision, and conversational interaction. We study the capabilities of state-of-the-art Large Vision Language Models (LVLMs) as overhearers to a corpus of spontaneous conversations between pairs of people engaged in a collaborative object-matching task. 

%OWEN PROPOSED REWRITE BELOW

During conversation, speakers collaborate on spontaneous referring expressions, which they can then re-use in subsequent conversation with the same partner. Understanding such referring expressions is an important ability for an embodied agent so that it can carry out tasks in the real world.  This requires integrating and understanding language, vision, and conversational interaction. We study the capabilities of seven state-of-the-art Large Vision Language Models (LVLMs) as overhearers to a corpus of spontaneous conversations between pairs of human discourse participants engaged in a collaborative object-matching task. We find that such a task remains challenging for current LVLMs, which fail to show a consistent performance improvement as they overhear more conversations from the same discourse participants repeating the same task for multiple rounds. We release our corpus and code\footnote{\url{https://github.com/jaaack-wang/lvlms-overhearing}} for reproducibility and to facilitate future research. 
% Our study also demonstrates that LVLMs are not robust enough to different object orderings nor to different people describing the same objects. 

\end{abstract}

\section{Introduction\label{sec:intro}}

% track: Multimodality and Language Grounding to Vision, Robotics and Beyond
A crucial skill for embodied AI agents working with humans is \textit{grounding in referential communication}:
%\susan{there are two senses of grounding that are relevant - an LVLM's ability to link words to images, and two interlocutors seeking and providing evidence to reach common grounding interactive conversation}\owen{I think this is now disambiguated, correct, Susan?}\susan{yes!} 
the ability to resolve which object in the visual environment a speaker is referring to.  
This is a non-trivial problem for several reasons: there may be no lexicalized label associated with the referent; there may be many ways to refer to it; or there may be multiple objects of the same type in the environment. Moreover, referential communication occurs in different interactive contexts: the referring expression can be part of a single, one-off instruction given to an AI agent; it can unfold over several conversational turns as a human interacts with the AI agent to clarify meaning; or the AI agent may overhear a conversation between two or more humans. The past few years have witnessed rapid advances in large vision language models (LVLMs) \citep[][\textit{inter alia}]{alayrac2022flamingovisuallanguagemodel, NEURIPS2023_6dcf277e, dai2023instructblipgeneralpurposevisionlanguagemodels, openai2024gpt4technicalreport, geminiteam2024geminifamilyhighlycapable}. However, LVLMs still lag behind human capabilities in both comprehending and generating referring expressions \cite{tang-etal-2024-grounding}, 
even in a single-instruction setting. 
%in a simple \textit{single-instruction} setting. 

% This raises the question about the potential of stat-of-the-art LVLMs deployed as an \textit{interactive}\owen{I find this confusing: while this is a good question, it is not the question we address in this paper.  I thin we should avoid the impression that we are dealing with interaction with AI agents, it will confuse reviewers.} embodied agent.\susan{I agree with Owen - this leads the reader to expect that this is about interaction. Stet please.}

\begin{figure}
    \centering
    \includegraphics[width=\linewidth]{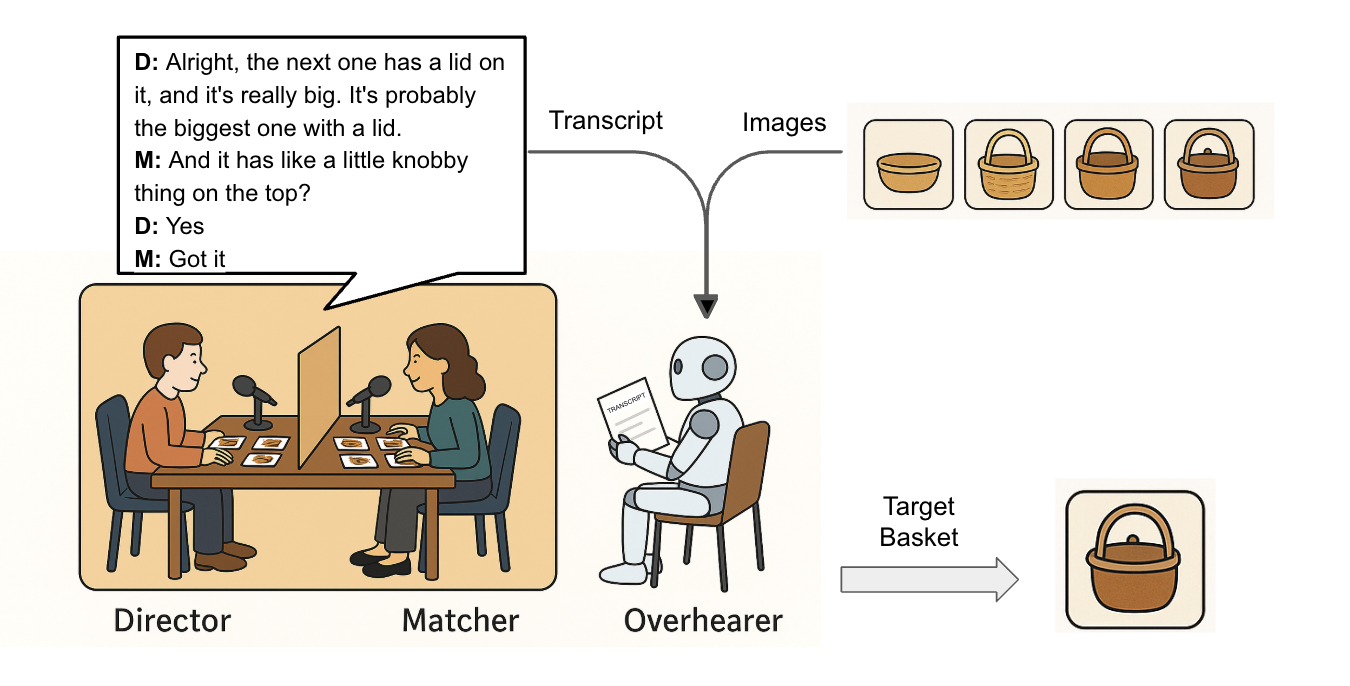}
    \caption{Our overhearer matching task (after \citealp[]{Schober1989}): the AI agent (LVLM) reads a transcript from a human referential communication corpus and tries to match the same cards as the matcher to the director's target sequence.}

    % \susan{please reformat ref as "(after Schober and Clark, 1989)" without the double parens}
    %\caption{The overhearer matching task in our study: the AI agent (LVLM) reads a transcribed spontaneous referential communication and tries to order the input objects in a correct sequence based on the conversation.}
    \label{fig:taskIllustration}
\end{figure}

In this paper, we address the problem of an AI agent overhearing two people engaged in spontaneous referential communication. 
%which remains underexplored. 
% \susan{lightly edited paragraph; redundancy removed.}
%(to the best of our knowledge) %Removed because we say the same thing more eloquently on the next page
This scenario is important 
%for at least two reasons: 
because the AI agent may be a side participant who stands by to assist when called on; it will need to 
understand the referential conventions that discourse participants develop over time. For example, assistive robots in the home may need to monitor conversations between residents (with prior consent) and perform tasks (such as manipulating objects) that require integrating and understanding language, vision, and conversational
interaction.

To address this issue, we use a previously unpublished corpus of spontaneous conversations between pairs of humans engaged in a collaborative object-matching task over repeated rounds of the same task. As illustrated in Figure~\ref{fig:taskIllustration}, we prompt an LVLM, acting as an overhearer, to perform the same task as the human matcher---that is, matching objects to the director's target sequence. 

We ask two research questions: (\textbf{1}) How well can LVLMs perform as overhearers in a referential communication task? and (\textbf{2}) Can LVLMs improve on their ability to resolve human-generated referring expressions, after witnessing repeated references to the same objects by the same human discourse participants? The second question is particularly important because even though psycholinguistics studies have shown that interacting in a conversation differs from overhearing it \cite{Schober1989,fox1999listening,tree2008overhearing,castano2023listening}, 
% \susan{lightly revised, as the "even though" logic of this sentence wasn't clear previously}
these two roles have not been defined or distinguished for LLMs. A useful AI agent in the overhearer role should adapt to and learn from the dynamics of language use, much like a human overhearer can. A failure to do so would suggest that the agent cannot effectively accumulate personalized knowledge across interactive 
%long 
contexts, thereby limiting its practical utility. To our knowledge, this study is the first to test LVLMs on a referent-matching task using a corpus of human referential communication. 
% \susan{I'm happy to report that we now say this only once!}

%\susan{I've cut the next sentence because it's repeated in items 2 and 3. Put it back in if you must.}
%Overall, our study indicates the limited practical utility of LVLMs as embodied agents in the context of spontaneous, interactive referential communication. 
\paragraph{}Our primary contributions are as follows:

\begin{enumerate}
    \item We release a corpus of spontaneous referential communication dialogues, collected under controlled conditions but previously unpublished, to facilitate future studies.
    %\susan{Will we need to add any conditions, such a password and a request to not include the corpus in training data?} \jack{Yes  I will do that. We can discuss it later.}
    
    \item We demonstrate empirically that resolving references to common real-world objects (i.e., baskets and dogs) produced during spontaneous conversation remains challenging for LVLMs, with high unexplained variability.
    
    \item We show that all tested LVLMs, including proprietary LVLMs like GPT-4o, fail to show a consistent performance improvement as they read more conversations from the same human pair %doing 
    matching the same 
    %matching task
    objects over time. 
\end{enumerate}

\section{Theoretical Background} 

% \jack{Susan/Weiling: feel free to elaborate this section where you see fit. Just remember that the camera ready has a 9-page limit for the main text (one more page than the submission).}
%\susan{Does the limit end with Conclusions, or does it include Acknowledgments and Limitations? if the latter, we're already over the limit...} \jack{It ends with Conclusions.}

Conversation, by its very nature, is collaborative. Speakers and addressees tailor their utterances 
and their interpretations to each other’s knowledge, needs, and perspectives, as well as to the \textit{common ground} they share \cite{clark1986referring}.  
%and extend through the conversation 
%\susan{cut because I don't know what this means!}
Sources of common ground can include co-membership in a community, as well as perceptual co-presence, which derives from interlocutors' mutual awareness of the shared environment. Most important for our purposes is linguistically-established common ground, or the prior co-presence of interlocutors to what they’ve said previously and can presume is part of their mutual knowledge \cite{clark1981definite}. The process of establishing and updating common ground is an essential engine for collaboration.
% \owen{This last sentence seems to not mention language use as the major course of CG that we are interested in, except "awareness of what they've said previously".  Also no mention of cooperating on CG.} 
%\susan{Owen, does this work better for you?}
%\weiling{modified based on Owen's suggestion. @Susan, please do the final check.}
%\owen{Yes, good}

\paragraph{Grounding}  The term \textit{grounding} has been used extensively in cognitive science, psycholinguistics, and AI. Grounding can be described as the ``access to or awareness of the physical, perceptual, goal-oriented or social contexts in which language occurs'' \cite{pavlick2023symbols},
%\owen{Is this a definition we like?  It's an NLP defnition and very broad.}, 
%\susan{Owen, I added this more specific definition, OK?}\owen{yes, good!}
and in human communication more specifically, the interactive process by which interlocutors seek and provide evidence that they understand one another, as they accrue common ground 
% \susan{Jack, please add a third ref here to: Brennan, S. E. (2005). How conversation is shaped by visual and spoken evidence.  In J. Trueswell & M. Tanenhaus (Eds.), Approaches to studying world-situated language use: Bridging the language-as-product and language-action traditions (pp. 95-129).  Cambridge, MA: MIT Press.}
\cite{clark1991grounding, METZING2003201, Brennan2005}.

% \susan{Jack, you commented my previous comment out without fixing the References list. Please provide the citation for Clark and Brennan - Clark, H. H., & Brennan, S. E. (1991).  Grounding in communication.  In L. B. Resnick, J. Levine, & S. D. Teasley (Eds.), Perspectives on socially shared cognition (pp. 127-149).  Washington, DC: APA.  Reprinted in R. M. Baecker (Ed.), Groupware and computer-supported cooperative work: Assisting human-human collaboration (pp. 222-233).  San Mateo, CA: Morgan Kaufman Publishers, Inc.}
%and add  bibtex for Metzing and Brennan 2003 (see text in code editor for more info)}

%Clark, H. H., & Brennan, S. E. (1991).  Grounding in communication.  In L. B. Resnick, J. Levine, & S. D. Teasley (Eds.), Perspectives on socially shared cognition (pp. 127-149).  Washington, DC: APA.  Reprinted in R. M. Baecker (Ed.), Groupware and computer-supported cooperative work: Assisting human-human collaboration (pp. 222-233).  San Mateo, CA: Morgan Kaufman Publishers, Inc.} 

% Metzing, C. & Brennan, S. E.  (2003).  When conceptual pacts are broken: Partner-specific effects in the comprehension of referring expressions.  Journal of Memory and Language, 49, 201-213.

This paper focuses on the grounding of linguistic expressions by pairs of people, mapped onto visual representations during a collaborative referential communication task.
%, specifically, ``the process by which speakers try to establish that what has been said is understood"
%\cite{clarkbrennan:1991}.
% Clark, H. H., & Brennan, S. E. (1991). Grounding in communication. In L. B. Resnick, J. M. Levine, & S. D. Teasley (Eds.), Perspectives on socially shared cognition (pp. 127–149). American Psychological Association. https://doi.org/10.1037/10096-006
%The notion of ``grounding'' relates a representation in one modality to one in another modality\owen{Owen will find citation for this}.  representation to a linguistic expression to its meaning. 
Although word meanings are informed by linguistic conventions,  meanings are not ``contained'' within words \cite{Reddy1993},
%\owen{This is basically good, but of course some part of meaning -- the lexical meaning --- is contained in words.  A "sofa" is not a "chair".  How about we change to: "Utterance meaning is not ``contained" in the words in the utterance, nor can it be retrieved from the lexical meanings of the words in the utterance using some simple procedure"}, but arise to some extent from convention, and in conversations,
%instead 
but can be collaboratively constructed by speakers and addresses, often in service of a shared task or goal (for discussion, see \citealp[]{brennan1996conceptual}).
%\susan{OK Owen? I revised to add conventional meaning.}\owen{Yes good}
 %\owen{Need a citation here, even if repeated later -- I assume BrennanClark} 
%Even strictly textual material (or at least that written by human authors) exists in a communicative context, oriented to an audience \cite{gerrig1993experiencing}. 
% \textbackslash{}cite\{Gerrig, 1993\}.
%Although LLMs are trained on vast amounts of data, that allow them to generate startlingly fluent text, they lack that one human partner's knowledge, needs, or perspective is distinct from another's. \jack{The last sentence: unless this is common knowledge or there are studies that support this conclusion, I think we should not say this directly, as it sounds like a speculation, or we should add a hedge here.}\owen{I agree with Jack}

\paragraph{Addressees vs Overhearers} 
% \susan{I swapped the order of these paragraphs and revised this section, as we need to explain performance of addressees before explaining performance of overhearers.} \jack{Looks good to me.}
Conversation is often studied in the lab using variants of a matching task, in which addressees interact
with speakers to match a set of picture
cards, build something together, or trace a route on a map.
%as in everyday communication outside of the lab, 
%\susan{changes removed b/c they didn't work with the rest of the sentence - "accuracy" isn't always the issue in everyday communication outside the lab. Keep this sentence about Schober \& Clark.}\owen{Ok (not sure who made the changes :) )}
The natural behavior of speakers and addressees in such tasks is to collaborate until they have evidence that they’ve reached a ``grounding criterion'' sufficient for their current purposes \cite{clark1991grounding, clark1986referring}; when the shared goal is to match a set of cards accurately, they continue seeking and providing evidence until they believe they've reached that criterion. The first round of a matching task always takes the longest (more time, words, and turns), becoming more efficient in subsequent rounds with the same objects and partners as they accrue common ground.

The role of an \textit{overhearer} is very different from that of an addressee. A now-classic psycholinguistics study by \citet{Schober1989} demonstrated that addressees in a matching task perform more accurately than overhearers in the same task, because they are able to ground meaning with speakers, whereas the overhearers cannot. 
%\weiling{I have smoothed it a bit.}
%(see Figure \ref{fig:taskIllustration}).
%The overhearers were at a disadvantage whether they were physically present from the beginning of the conversation, or just heard a recording of the interaction later on. 
%In contrast to an addressee, a
An overhearer sometimes understands the referent \textit{early} (but must wait for the task to move on), sometimes \textit{late} (and falls behind in the task) and sometimes not at all (and selects the wrong picture card). In \citet{Schober1989}, addressees (who could contribute to the conversation) matched the cards nearly perfectly,
%accurately 93\% of the time in Round 1 and 100\% of the time in all later rounds, 
whereas overhearers (who heard every word but did not interact)
%the director and matcher said (either from a recording in Study 1 or from being present in the room from the start in Study 2) 
reached only about 80\% accuracy in Round 1 and about 90\% by Round 4. Strikingly, ``late overhearers'' who listened in to the recorded conversations starting in Round 3 did worst of all, achieving only 68-73\% accuracy, even when they could stop and start the recordings to try to keep up  (See Figure 3 in their paper).

\paragraph{Variation in Human Language Use}  
There is considerable variation in human language use, including in choices of wording, syntax, prosody, and coordination strategies. This variation is not random, but emerges strategically, such that lexical variability is much greater across conversations than within a conversation. For instance, two partners in dialogue tend to \textit{entrain} on words, consistently using the same referring expression (albeit in shortened form) over the course of repeated referring to the same object, as if to confirm a ``conceptual pact'' that they're referring to the same thing they discussed previously \cite{brennan1996conceptual, Krauss1964, METZING2003201}.

\paragraph{Challenges for an AI Agent} Ultimately, a useful embodied AI agent should be able to maintain, adapt, and build on 
%representations of 
the common ground accrued in conversation.
%\owen{"representations of CG" is odd -- "referential conventions which are in the CG"?} \weiling{I feel it's better to just say "maintain, adapt, and use the common ground accrued in conversation"}  
The agent should (1) be robust enough to understand and track the expressions that different pairs of speakers use to describe the same objects \cite{brennan1996conceptual}, and (2) be able to cope with the dynamic variations in human language use, including the choice of wording, syntax, prosody, and coordination strategies (especially those due to the use of common ground).
%when human partners achieve more efficiency upon repeated referring within the same environment.\owen{I am not 100\% certain what "dynamic and variable nature of langauge use" means, super general -- make this more precise?} \weiling{Modified based on Owen's suggestion.} 
This requires evaluating how well a foundation model (whether LLM or LVLM) performs with spontaneous 
%and dynamically evolving 
dialog during repeated discussions of the same objects with the same speakers.
%rarely studied (if at all).
%\susan{I removed "rarely studied, as we say this multiple times (e.g., "unexplored", etc.}\owen{Ok}

%\paragraph{AI Agents as Overhearers}  Ultimately, a useful embodied AI agent should be able to maintain, adapt to, and use representations of common ground accrued in conversation. It should be able to cope with the considerable variation in human language use, including in choices of wording, syntax, prosody, and coordination strategies. This variation is not random. For instance, two partners in dialogue tend to entrain on and consistently use the same referring expression over the course of repeated referring to the same object, such that lexical variability is much greater across conversations than within a conversation \cite{BrennanClark1996}. And as pairs in referential communication studies match the same set of objects in repeated rounds, their referring expressions become much more efficient as they accrue common ground. This means that an AI agent should (1) be robust enough to understand that different pairs of human interlocutors use different expressions to describe the same visual object \cite{BrennanClark1996} \cite{others}, and (2) be able to capture the dynamics of spontaneous language use, as human partners achieve more efficiency upon repeated referring within the same environment. However, evaluating how well a foundation model (whether LLM or LVLM) performs under spontaneous and dynamically evolving dialogue across repeated runs of the same task is barely studied, if at all.

\section{Related Work} 

\paragraph{Machine Comprehension of Referring Expressions} %A common approach on evaluating the visual grounding capabilities of language models
% \susan{is this phrase ok Jack? A wee transition was needed here.} \jack{Panos, can you double check? I think this is a very general statement and not necessarily related to language models. REC has existed before LMs come along.}
%involves identifying a specific object in an image based on a natural language description \cite{rec-survey}. 
Evaluating the visual grounding abilities of language models often involves tasks that require identifying a particular object in an image using a natural language referring expression \cite{rec-survey}. Conventional methods typically follow a two-stage approach: first generating open-vocabulary object proposals, then selecting the one that best matches the language description. More recent efforts have built upon the emerging capabilities of vision transformers  
\cite{dosovitskiy2020image}, leading to improved models \cite{su2024scanformer}. Moreover, LVLMs have demonstrated strong performance on visual grounding tasks even in zero-shot settings \cite{sui2023language, subramanian-etal-2022-reclip}, while specialized approaches have been developed to further improve their zero-shot grounding abilities without task-specific supervision \cite{han2024zero}.

%\paragraph{Visual Grounding of LVLMs}
% \susan{great section, Panos!}
% \susan{I've swapped the order of Panos' section and the Tang paragraph}
% \owen{We need to talk about ToM work somewhere, since it is overhearing and we mention ToM}
% Mockup structure of this subsection:
% \begin{itemize}
% \item Quick intro of the emergence and the capabilities of LVLM's to perform visual grounding
% \item Key limitations of LVLMs in visual grounding.
% \item Recent advances
% \item Multi-Turn and Dialogue-Based Multimodal Grounding
% \end{itemize}

\paragraph{Visual Grounding Capabilities of LVLMs} LVLMs exhibit strong visual grounding capabilities thanks to their large-scale multi-modal pretraining \cite{sem-grounding}. On top of the core vision-language alignment principles established by foundational models, such as CLIP \cite{clip}, LVLMs show remarkable performance on a wide range of tasks, including referring expression comprehension \cite{ref-comprehension}, visual question answering \cite{vqa-vlms}, and instruction following \cite{bitton2023visitbench}. However, spatial and compositional reasoning remains a challenging task for current LVLMs, as, for example, they often struggle with relational cues \cite{Chen_2024_CVPR} or multiple visual concepts \cite{compositional-reasoning-challenges}.

Another line of research explores the interactive visual grounding capabilities of LVLMs engaged in multi-turn dialogues to resolve ambiguous references and refine their understanding through conversational context \cite{feng-etal-2023-mmdialog, tian2025chatterbox}. Recently, \citet{tang-etal-2024-grounding} showed that humans consistently outperform LVLMs in comprehending referring expressions generated to target a dot placed in a shared 3-D environment. Our study extends this work to focus on the capabilities of LVLMs as overhearers, to examine the mapping of referring expressions onto referents discussed repeatedly in spontaneous human conversations, rather than as one-off descriptions.

Finally, although LVLMs may be able to process %increasingly efficient 
language produced by humans, they do not exhibit a human-like tendency to spontaneously make the language they generate more efficient over multiple turns \cite{hua2024talk}.

%\jack{This may not be neccessary as our paper does not focus on referring expression generation. We do not need a separate paragraph discussing this paper.}

% \paragraph{Performance of LVLMs with Humans} In parallel, other approaches explore interactive visual grounding, where models engage in multi-turn dialogues to resolve ambiguous references and refine their understanding through conversational context \cite{feng-etal-2023-mmdialog, tian2025chatterbox}. \citet{tang-etal-2024-grounding}\owen{I think this fits with the last part of Panos' text above, we don't need a new subsection.  I added a para break in Panos's text, this can just be another para} investigated how well an LVLM-based agent can generate referring expressions for another agent who was co-present in the same spatial scene, but at a different point in the scene than the LVLM. The finding was that GPT-4o could generate expressions (about the spatial location of a dot) that were comprehended successfully by humans with a mean of 64.9\%, compared to a mean of 87.6\% success for expressions generated by humans. However, in that study, there was no interaction permitted, nor did the expressions map to realistic objects that humans typically refer to in the real world. Our project aims to quantify the current ability of LVLMs to understand the grounding of referring expressions to visual objects in overheard interactive human dialogue. 
% In this paper, we examine the capabilities of LVLMs to overhear such multi-turn dialogues and understand common ground reached between the two human subjects. 

\begin{figure*}
    \centering
    \includegraphics[width=0.8\linewidth]{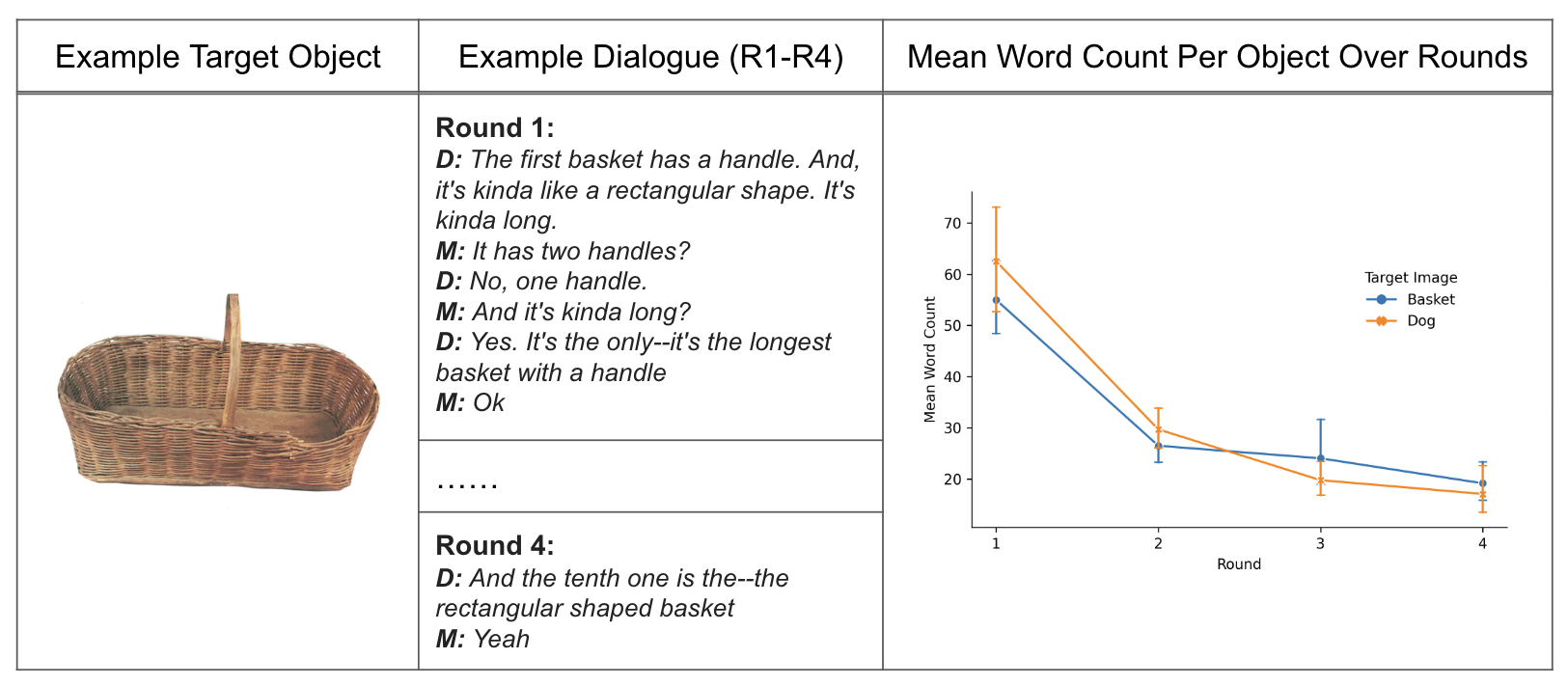}
    \caption{The left panel shows one target basket; the middle panel shows one pair's corresponding dialogue from Round 1 to Round 4, demonstrating entrainment on more concise language (for the perspective "rectangular-shaped"). Here, entrainment occurs after they consider multiple proposals in Round 1. The right panel depicts the mean word count (a measure of efficiency) for baskets and dogs across rounds. Error bars indicate $\pm$1 standard error of the mean across pairs.}
    \label{fig:corpus_example}
\end{figure*}

\paragraph{Studies of Machines as Overhearers} Numerous studies have evaluated LLM/LVLM foundation models in the overhearer/observer role \cite{castano2023listening, kim-etal-2023-fantom, Kosinski2024, jin-etal-2024-mmtom, soubki-etal-2024-views, street2024llmsachieveadulthuman}. Most 
%of them 
pertain to Theory of Mind (ToM) \cite{Premack1978} and examine a model's ToM ability to attribute false beliefs to characters due to their absence at a critical point in the story, following the classic Sally-Anne test \cite{Wimmer1983, BaronCohen1985}. Our study differs from ToM studies in that we analyze the grounding capabilities of LVLMs without assuming information asymmetry between the overhearer and the discourse participants: all 
have access to the same words, as in \citet{Schober1989}.
%heard as much as each other.\susan{minor edit to polish this.}

% Moreover, our evaluations are based on spontaneous human conversations, instead of typically template-generated stories. \jack{I guess this is not the main point.}

\section{Corpus\label{sec:corpus}} %\jack{Figure~\ref{fig:corpus_example} is not  referred to in this section at all.}

\paragraph{Overview}
%\jack{Susan/Weiling: Details about corpus can now be revealed, e.g., where and when was the corpus created. Please add.} 
Our corpus comprises 80 human-to-human dialogues totalling 27,902 words, collected by Calion B. Lockridge and Susan E. Brennan at Stony Brook University in 2001 and not previously published. Ten pairs of native-English-speaking undergraduates (20 speakers in total) did repeated rounds of a referential communication task \cite{krauss1969development, clark1986referring}. During each round, the pairs spoke freely 
while they matched duplicate sets of picture cards. The dialogues were recorded and manually transcribed. Figure
%~\ref{fig:taskIllustration} and 
\ref{fig:corpus_example} shows a representative example excerpted from the transcripts of two people describing the same target basket in Round 1 and again in Round 4.

\paragraph{Task and Materials} Following \citet{clark1986referring}, %twenty undergraduate native speakers of English  — none of whom had met or interacted previously—
speakers were recruited in pairs, with one partner randomly assigned to the role of director (D) and the other to the role of matcher (M); they remained in their assigned role throughout the experiment. Partners sat in separate rooms and communicated via an audio channel.

Each pair completed a total of eight rounds of the referential communication task in a one-hour session---four rounds with the same set of pictures of dogs,
%\owen{Were these teh same day? How much time had elapsed?}\susan{immediately afterward. The whole experiment took about an hour. Note that I added "the same" to "set", to address your other question}, 
and four rounds with the same set of pictures of baskets (counterbalanced for order). 
These basic-level categories were chosen %intentionally 
to vary the difficulty of expressing and identifying referents
%for a meaningful evaluation
: dogs are associated with commonly-known subordinate category labels such as breeds, whereas baskets are not (see Figures~\ref{fig:baskets} and ~\ref{fig:dogs} in Appendix~\ref{app:corpus} for details). D's and M's sets contained duplicates of the same 10 dogs (or baskets), with 3 additional cards only in M's set,
%that D did not have, 
to require them to discuss all 10 targets. 
%For each of baskets and dogs, in each of the four rounds, the D and M got exactly the same set of cards, so that the tasks were identical.\owen{Weiling/Susan: Is the preceding sentence correct?}
%\susan{Owen, I think this is clear from the first sentence, but I added "the same" to "set" just in case.}

%introduce variations in difficulty of expressing and identifying referents: dog breeds offer rich, nameable distinctions, while baskets are more abstract and harder to describe. See task images in Appendix \ref{app:corpus}. Task order (dogs-first vs. baskets-first) was counterbalanced across pairs. 

%In each round, participants were seated across a physical barrier to prevent visual contact and relied entirely on verbal communication. The director described a sequence of 10 target images, and the matcher selected corresponding cards from a larger set of 13, aiming to complete the sequence as accurately and efficiently as possible.

\paragraph{Performance and Linguistic Patterns}
%\weiling{I modified the section title}
%Performance and Summary Statistics}
All pairs successfully completed the matching task in all rounds, achieving 100\% accuracy. 
%On average, each pair used 38.60 words per4 round to describe each basket and 31.16 words to describe each dog. 
This 
%The Lockridge and Brennan
corpus showed consistent linguistic patterns that align with the findings of \citet{clark1986referring} and \citet{Schober1989}, with partners becoming more efficient in their expressions across rounds. %\owen{Susan: who is "we"?  The reader will assume that "we" is the authors of tihs paper, is that as intended?  If you mean Lockridge and yourself, need to say} 
%and using fewer words in later rounds\owen{Added explanation of "efficient" -- ok?} ocr: I see there is  a whole para
That is, objects were described in greatest detail in Round 1, often with multiple proposals for expressions until M acknowledged understanding.
%averaging 59.3 words for dogs and 58.8 words for baskets). 
By as early as Round 2, word counts dropped sharply, by about 50\%.
%- to 28.7 words for dogs and 29.2 for baskets. 
By Rounds 3 and 4, partners typically had entrained on shared conceptualizations, with concise labels for the objects 
%\owen{Susan: I am not sure what "perspectives" means here -- just keep "labels"?-- possibly ambiguous to readers with respect to visual perspective} and labels for the same objects.
%, averaging just 17.2 words for dogs and 22.1 words for baskets. 
This reflects the accumulation of common ground over repeated interactions. The summary plot in Figure~\ref{fig:corpus_example} illustrates this in the form of 
%trend, through both a representative dialogue example and a summary plot showing 
reduced word counts across rounds (see also Figure \ref{fig:word count changes per role} in Appendix~\ref{app:corpus}).

\paragraph{Manual Extraction of Object Descriptions}
%\susan{Remind me how/why we use these? to essentially makes it easier for the model, yes? If so, can we add a statement here like "To conduct followup studies...", and should this be moved to where we describe those studies?} 
% \jack{We use this in one of the experiments in Section~\ref{sec:follow-up}, but the extracted object descriptions can also add a value to the corpus, because, as said in the last sentence, it allows for a finer examination of a system's visual grounding capability on an object level}
% \susan{Locating this paragraph here could imply that the LVLMs had this help for all upcoming expts. But see if the text I added here makes it clear enough (alternatively, we might move this to Section 7 where we do the follow-up expt).}
Our experiments use the transcripts from these spontaneous conversations; however, we extended the corpus for a follow-up experiment (see the Object Descriptions test reported in Section~\ref{sec:follow-up}) by manually extracting 10 complete object descriptions from each transcript, yielding a 800 object descriptions. Each description starts with D describing a target object and ends when M recognizes it. This allows for a finer examination of a system's visual grounding capability on an object level.

% \jack{Added this paragraph.} \weiling{slightly modified it}

\paragraph{Corpus Value and Lack of Data Contamination} 
%Since the corpus was created two decades before the release of ChatGPT and has not been published, it ocr: again, let' skeep details about teh corpus that could help identify us authors out of the paper
As this corpus has not been published and is not included in any LVLM training data, it is free from the risk of data contamination \cite{jacovi2023stop, sainz2023nlp}. It provides an ideal testbed for evaluating LVLMs' ability to adapt to spontaneously produced referring expressions from multiple speakers, grounded in visual images, without the influence of prior exposure or memorization.

\section{Methodology\label{sec:methodology}}

% This section describes our research hypotheses and the experiments designed to test these hypotheses. 

% \subsection{Research Hypotheses}

% \label{sec:rh}

% \newcommand{\rhone}[0]{\textbf{RH1}}
% \newcommand{\rhtwo}[0]{\textbf{RH2}}
% \newcommand{\rhthree}[0]{\textbf{RH3}}

% We summarize\owen{Jack: you are suggesting removing this subsection, correct?} our research hypotheses, which are based on the (perhaps wrong) assumption that LVLMs will behave like humans. \jack{For each hypothesis, we should have one sentence justify it in connection to previous cognitive science results.}

% \noindent {\bf Increased performance over rounds for a fixed start}: the results get better with increasing rounds, no matter which round the LVLM starts following the conversation (\rhone{}).  

% \noindent {\bf Decreased performance by starting round}: the later the LVLM starts following the conversation, the worse the performance for its starting round (\rhtwo{}).

% \noindent {\bf Irrecoverable performance loss for late start}: the later the round in which the LVLM starts following the conversation, the worse the performance for the fourth and final round (\rhthree{}).

\begin{figure*}
    \centering
    \includegraphics[width=\linewidth]{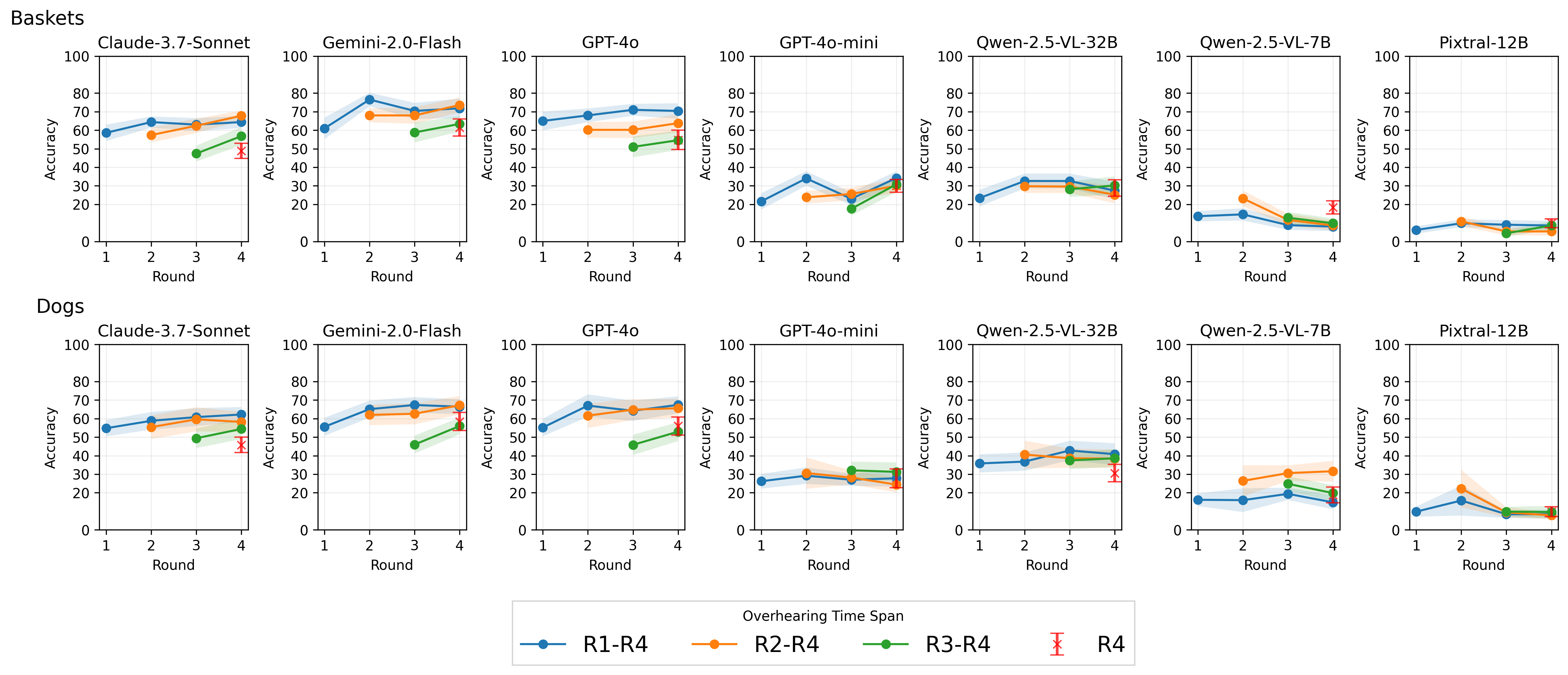}
    \caption{Average accuracy of various LVLMs in the overhearer task over rounds. There are 4 overhearing starting points from Round 1 to Round 4, yielding three lines and one single point. The shaded areas and error bars denote 95\% confidence intervals. In this corpus, all human matchers' performance is 100\% at every round.}
    \label{fig:main_results}
\end{figure*}

\paragraph{Task Description} To prompt an LVLM to perform the overhearer matching task, we provide it with a transcript and an image as inputs. The transcript is a conversation between a director and a matcher at a specific round from our corpus. The input image contains the corresponding 13 objects (baskets or dogs) used during the conversation, randomly arranged in a 3x5 grid, and numbered from 1 to 13 (see Figures~\ref{fig:baskets} and ~\ref{fig:dogs} in Appendix~\ref{app:corpus} for two examples). The LVLM is instructed to produce the correct sequence of 10 target object indices for the objects as described by the director.

\paragraph{Experimental Procedure} Our corpus contains four rounds of conversations between each human director-matcher pair for each object type. To address the two research questions listed in Section~\ref{sec:intro}, 
%regarding the capabilities of LVLMs resolving references and adapting to the dynamics of language use\jack{Please rewrite.}, 
we evaluate (1) the performance of an LVLM as overhearer at single rounds and (2) how this performance evolves over multiple rounds of conversation between each director-matcher pair, with the methods and performance of \citet{Schober1989}'s overhearers in mind. 
More concretely, we measure LVLM task performance in each round from a starting round to the end round.  We tested four starting rounds for each object type, i.e., Round 1 (R1), Round 2 (R2), Round 3 (R3), and Round 4 (R4), and one end round, R4. We prompt the LVLM to perform the matching task for each starting round separately
% \owen{Jack: added this word, correct?  Found it hard o understand without} 
in a multi-turn conversation setting. 

% The transcripts for each series of rounds are provided in the order they occurred during the human experiments
% \owen{Jack: What does this mean -- the order of the human pairs is as in the corpus?  Or the order of the rounds is as in the corpus?  Please clarify.}. \jack{Owen, better now?}

The input objects are shuffled for each round, since the human overhearer's display showed the objects in arbitrary order.
% \owen{So we are saying it is not the same as the M saw, right?  Because we don't know anymore what the M saw.  This sentence seems to make a distinction between the M and overhearer, but that is not relevant.} 
% \susan{the human matchers also saw their cards arranged in an arbitrary order, different for each round}
Our early experiments showed that LVLMs appear to be sensitive to object orderings in their visual input,
% \owen{I feel we should also acknowledge just random variation in LVLM output as well, which is another reason to do 5 experiments} \jack{That will require too much writing}
which we further evaluate in Section~\ref{sec:robustness}. While this %empirical observation 
sensitivity to ordering reflects a challenging aspect of our corpus, offering an ideal testbed for robustness testing, it poses an evaluation challenge. To minimize effects of order and obtain more reliable results, we run each LVLM five times with different object orderings via greedy decoding for each experimental configuration throughout the study.

\paragraph{LVLMs} We evaluate (\textbf{1}) four proprietary LVLMs, namely Claude-3.7-Sonnet \cite{claude3.7sonnet}, Gemini-2.0-Flash \cite{google2024gemini2}, and GPT-4o and GPT-4o-mini \cite{openai2024gpt4ocard}, as well as (\textbf{2}) three open-weight LVLMs, namely Qwen2.5-VL-32B and Qwen2.5-VL-7B \cite{Qwen2.5-VL}, and Pixtral-12B \cite{agrawal2024pixtral12b}. We choose only models that support multiple input images, since our task involves multiple input images from different rounds. See Appendix~\ref{app:models} for more details about these models.

\paragraph{Prompting} In the prompt, we provide all the necessary background information regarding the human director-matcher matching task, as described in Section~\ref{sec:corpus}. We explain the overhearer matching task and the task procedure, as illustrated earlier. We prompt all LVLMs with zero-shot chain-of-thought \cite{zeroshotCoT}, with temperature set to 0 to maximize reproducibility. All the prompt templates can be found in Appendix~\ref{app:prompts}. 

% \owen{Jack: For clarity, say in the caption of this figure that the bottom part is repeated up to 4 times in a conversation before each transcript/image pair -- that is correct, no?} \jack{Which figure? If you meant Figure~\ref{fig:main_results}, the caption already says that there are from starting rounds from R1 to R4. Isn't it clear?}

\paragraph{Evaluation Metric} We compute accuracy, i.e., percentage of correctly matched objects, to measure LVLM task performance on the overhearer matching task. In other words, a model getting 9/10 or 0/10 objects correct would score 90\% and 0\%, respectively. 
%\jack{Added this sentence for clarity.}

\section{Results\label{sec:results}}

Figure~\ref{fig:main_results} shows the average accuracy (with 95\% confidence intervals) of various LVLMs on the overhearer matching task across multiple rounds. We compare performance among the models for a single round and for sequences of rounds with different starting  points (capturing the experience of \citet{Schober1989}'s ``late overhearers'').

To interpret these results, we first examine the performance of the LVLMs at single rounds to evaluate how well they 
resolve real-world object references in spontaneous, interactive conversations. We then analyze how their performance evolves when initialized from different starting rounds, providing insights into their potential as embodied agents. Lastly, we conduct a robustness analysis.

\subsection{Performance at Single Rounds}

\paragraph{Proprietary LVLMs substantially outperform open-weight models.} As Figure 3 shows, the large proprietary LVLMs (i.e. GPT-4o, Gemini-2.0-Flash, and Claude-3.7-Sonnet) achieve an average accuracy ranging from 45.8\%\ (Claude-3.7-Sonnet at R4 with a R4 start for dogs) to 76.5\% (Gemini-2.0-Flash at R2 with a R1 start for baskets). In contrast, the open-weight models achieve only 42.8\% accuracy at best (Qwen-2.5-VL-32B at R3 with a R1 start for dogs), and 4.4\% at worst (Pixtral-12B at R3 with a R3 start for baskets).
%best-performing open-weight model, Qwen-2.5-VL-32B, achieves only 42.8\% accuracy (at R3 with a R1 start for dogs), while the worst, Pixtral-12B, is only 4.4\% (at R3 with a R3 start for baskets).
%\owen{I think we need to give more consistent numbers, this is too impressionistic.  Maybe minimum and maximum delta between the two model types?} \jack{I think this is enough. The figure is quite self explanatory.}
%\susan{I've rephrased slightly and pointed to the figure to make it less impressionistic -- does this work Owen?}\owen{Yes good}

\paragraph{Model scaling appears beneficial for language grounding.} 
%It is unclear if model scaling contributes to the superior performance observed in the , as 
Model sizes of proprietary LVLMs are not publicly disclosed. But available information suggests that size and performance are correlated. Specifically, GPT-4o consistently outperforms GPT-4o-mini, and Qwen-2.5-VL-32B outperforms Qwen-2.5-VL-7B.
%\owen{Is this significant?}\jack{This is very obvious. I will do the test if i have time.}.  
Furthermore, proprietary LVLMs are likely larger than other models, suggesting that larger models perform better.

\paragraph{LVLMs underperform human matchers.} Recall that in our corpus, all human director-matcher pairs completed the matching task with 100\% accuracy in every round, substantially outperforming all tested LVLMs, regardless of LVLM starting round. Prior research has shown that human overhearers also perform worse than interacting partners in matching tasks \citep{Schober1989, fox1999listening, tree2008overhearing, wilkes1992coordinating}. %castano2023listening}.
% \jack{This is the second time I repeat this sequence of references. Susan, rewrite this sentence if needed.} 
% \susan{Sentence lookd good. Castano didn't use a matching task but a ToM task, so I've moved that ref from here to the Studies of Machines as Overhearers section, as we don't cite it anywhere else.}
But unlike human overhearers, LVLM overhearers in our experiments can access the entire conversation for every matching decision, thanks to their built-in attention mechanisms. Despite this advantage, even state-of-the-art LVLMs fail to reliably exploit this additional conversational context to match human-level performance. This suggests that \textit{grounding spontaneous, naturalistic descriptions to visual referents remains a substantial challenge for LVLMs.}

\begin{table}[]
    \centering
    \scriptsize
    
\begin{tabular}{lllll}
\toprule
 & Starting Round & R1 & R2 & R3 \\
Source & Model &  &  &  \\
\midrule
\multirow[t]{7}{*}{Baskets} & Claude-3.7-Sonnet & 1.6 & {\textcolor{ForestGreen}{5.2***}} & {\textcolor{ForestGreen}{9.4**}} \\
 & Gemini-2.0-Flash & {\textcolor{ForestGreen}{2.6*}} & 2.8 & 4.6 \\
 & GPT-4o & {\textcolor{ForestGreen}{1.9*}} & 1.8 & 3.6 \\
 & GPT-4o-mini & {\textcolor{ForestGreen}{2.7**}} & {\textcolor{ForestGreen}{3.1*}} & {\textcolor{ForestGreen}{13.2***}} \\
 & Qwen-2.5-VL-32B & 1.2 & -2.3 & 2.0 \\
 & Qwen-2.5-VL-7B & {\textcolor{red}{-2.3***}} & {\textcolor{red}{-7.3***}} & -3.0 \\
 & Pixtral-12B & 0.6 & {\textcolor{red}{-2.7**}} & {\textcolor{ForestGreen}{4.2**}} \\

\midrule \midrule

\multirow[t]{7}{*}{Dogs} & Claude-3.7-Sonnet & {\textcolor{ForestGreen}{2.4*}} & 1.4 & 5.0 \\
 & Gemini-2.0-Flash & {\textcolor{ForestGreen}{3.5***}} & 2.6 & {\textcolor{ForestGreen}{10.0**}} \\
 & GPT-4o & {\textcolor{ForestGreen}{3.4**}} & 2.0 & 7.1 \\
 & GPT-4o-mini & 0.3 & -3.1 & -0.8 \\
 & Qwen-2.5-VL-32B & 2.1 & -1.1 & 1.2 \\
 & Qwen-2.5-VL-7B & -0.1 & 2.6 & -5.0 \\
 & Pixtral-12B & -1.2 & {\textcolor{red}{-7.2**}} & -0.2 \\
\bottomrule
\end{tabular}

% old caption
%\caption{Overall performance trend (slope) over rounds for each LVLM starting at R$_i$, using ordinary least squares (OLS) regression. We highlight both \textbf{significant} \textcolor{ForestGreen}{positive} and \textcolor{red}{negative} trends; we indicate significance using asterisks: ``*'' means \textit{p} < 0.05, ``**'' means \textit{p} < 0.01, and ``***'' means \textit{p} < 0.001.}

    \caption{Overall performance trend (slope) over rounds for each LVLM starting at R$_i$, using ordinary least squares (OLS) regression. Significant 
    %\textbf{Significant} 
    \textcolor{ForestGreen}{positive} and \textcolor{red}{negative} trends are highlighted, along with  significance (``*'' is \textit{p} < 0.05, ``**'' is \textit{p} < 0.01, and ``***'' is \textit{p} < 0.001.) See Figure~\ref{fig:main_results} for a visualization of the performance trend.}
    \label{tab:performanceOverRounds}
\end{table}

\subsection{Performance Dynamics Across Rounds}

% Humans are known to be able to improve as overhearers as they listen to the same two participants doing the same matching task over rounds, even when the human overhearers missed the first %few 
% two rounds \cite{Schober1989}. That said, the timing of when overhearing begins also matters: late overhearers who missed the buildup of common ground between conversational participants in the early rounds understand fewer references in the later rounds and are thus found to consistently underperform earlier overhearers.\susan{this is because the bulk of the work leading to entrainment occurs in the first couple of rounds.} We show below that LVLMs generally do \textit{not} exhibit such behaviors.

\paragraph{LVLMs fail to show consistent improvement 
on the same matching task over time.} 
To measure how each model's performance evolves over time, we use ordinary least squares (OLS) regression to model the overall performance trend shown in Figure~\ref{fig:main_results}. A run shows improvement only if the regression line has a positive coefficient with a \textit{p}-value less than $0.05$. The results in Table~\ref{tab:performanceOverRounds} confirm the performance gap between proprietary LVLMs and open-weight LVLMs, with the former  showing more desirable performance trends. However, even for the proprietary LVLMs, overall performance does not consistently improve for every starting round (the only exception is, surprisingly, the smallest proprietary model, GPT-4o-mini, 
for baskets). Often, the open-weight models even decrease.
%Figure~\ref{fig:main_results} shows that its performance fluctuates greatly (e.g., plummets at R3) when it starts at R1.\owen{So what?  Linear regression is highly significant.  This is only one of three.  This is weak -- just acknowledge the exception with no "however"?} 
We also analyze the overall performance trend using Spearman and Kendall rank correlations, which further validate these findings and show that even for a positive performance trend, the overall correlations remain small (see Appendix~\ref{app:overallPerformanceTrend}).  

We further measure the performance trend for the dialogues of each human pair, using  the same method. We then compute the percentage of human pairs for whom the LVLMs show consistent improvement over the  starting rounds. 
% \susan{mea culpa - I added this. Figure 4 is the only figure that breaks out performance of 2 LVLMs on each human pair. Therefore I'm confused too. The "see also Table 7" suggests that we should also refer to this result in another figure that's not in the Appendix.} \jack{Susan, there is a misunderstandings. Figure 4 does not show LVLM performance on each human pair ACROSS rounds. There, we simply run each LVLM on each human pair 30 times to evaluate LVLM performance robustness.}
The results show that the models tested fail to show an overall improvement on at least 70\% of the human pairs (see also Table~\ref{tab:performanceOverTimeAcrossHumanPairs} in Appendix~\ref{app:overallPerformanceTrend}). 

Lastly, as a sanity check, we simply compute the percentage of times an LVLM shows monotonically increasing 
%(or non-decreasing) 
performance for each starting round across the two datasets. The results show that all LVLMs struggle to achieve a smooth, incremental performance improvement over time, since, for example, when starting from R1, the best model exhibits a monotonically increasing performance curve only 46\% of the time. See Table~\ref{tab:performanceMonotonicity} in Appendix~\ref{app:overallPerformanceTrend} for details.

\begin{table}[]
    \centering
    \small 

% \begin{tabular}{l|cc|cc}
% \toprule
%  & \multicolumn{2}{c}{\% Sig. Pos} & \multicolumn{2}{c}{\% Sig. Neg} \\
% Source & Baskets & Dogs & Baskets & Dogs \\
% Model &  &  &  &  \\
% \midrule
% GPT-4o & 0 & 0 & 30 & 0 \\
% GPT-4o-mini & 10 & 10 & 0 & 0 \\
% Gemini-2.0-Flash & 10 & 0 & 20 & 0 \\
% Claude-3.7-Sonnet & 0 & 0 & 40 & 20 \\
% Pixtral-12B & 0 & 0 & 0 & 0 \\
% Qwen-2.5-VL-7B & 10 & 10 & 0 & 0 \\
% Qwen-2.5-VL-32B & 20 & 10 & 10 & 10 \\
% \bottomrule
% \end{tabular}
    
%     \caption{Percentage of time an LVLM's performance across the four starting rounds shows a significant positive trend and a significant negative trend.}

\begin{tabular}{lll}
\toprule
Source & Baskets & Dogs \\
Model &  &  \\
\midrule
Claude-3.7-Sonnet & {\textcolor{red}{-3.9***}} & {\textcolor{red}{-3.3**}} \\
Gemini-2.0-Flash & -0.8 & -0.8 \\
GPT-4o & {\textcolor{red}{-4.0***}} & -1.3 \\
GPT-4o-mini & {\textcolor{ForestGreen}{1.9*}} & 0.6 \\
Qwen-2.5-VL-32B & 1.5 & -1.9 \\
Qwen-2.5-VL-7B & 0.4 & 0.6 \\
Pixtral-12B & 0.4 & -1.2 \\

\bottomrule
\end{tabular}
\caption{Overall performance trend (slope) across the four starting points (R1, R2, R3, and R4) for each LVLM, using ordinary least squares (OLS) regression. A significant negative slope is more desired here as that indicates that an model benefits from an earlier start.}

    \label{tab:performanceOverStartingRounds}
\end{table}

\paragraph{LVLMs do not consistently benefit from an early start, unlike humans.} Given the characteristics of our corpus where object descriptions used in R1 tend to be much more elaborate than the subsequent rounds (see Figure~\ref{fig:corpus_example}), we expected LVLMs to perform better in their starting round if they begin with R1 than with R2 (or other later rounds).  However, no models 
show a significantly better performance in R1 than R2 as their starting rounds in paired t-tests (see Table~\ref{tab:startingRoundsDifsTTestsBaskets} in Appendix~\ref{app:results}). In fact, several LVLMs perform significantly better in their starting round when beginning with R2 rather than R1, including Gemini-2.0-Flash for baskets (7\% mean accuracy gain for R2, $p < 0.05$.) 
%\jack{We do not have to mention other pairs of starting rounds, such as R2 versus R3. This demonstrates the point.}

To study the overall performance differences between starting at an earlier round versus a later round, we use OLS regression to analyze the overall performance trend across the four starting 
%rounds 
points (i.e., R1, R2, R3, and R4). Table~\ref{tab:performanceOverStartingRounds} shows that two proprietary LVLMs (GPT-4o and Claude-3.7-Sonnet) generally perform better the earlier overhearing starts, but none of the 
%open-weight 
other LVLMs show such a performance pattern. 

In principle, with an earlier start, an LVLM reads in more conversation between the human director and matcher, so it should better understand the subsequent rounds of conversations, compared to when it begins at a later round. We compare the mean accuracy differences of overlapping rounds between an early start and a late start using paired t-tests 
%and report the results in 
(see Table~\ref{tab:meanDiffBetweenOverlappingRounds} in Appendix~\ref{app:results}).
%for space reasons. 
%We find that p
Proprietary LVLMs tend to benefit more from an early start, in terms of performance gain over subsequent rounds, than open-weight models. This suggests that these open-weight models are less able to use previous discourse effectively to capture the dynamics of language use over time.

\begin{figure}
    \centering
    \includegraphics[width=\linewidth]{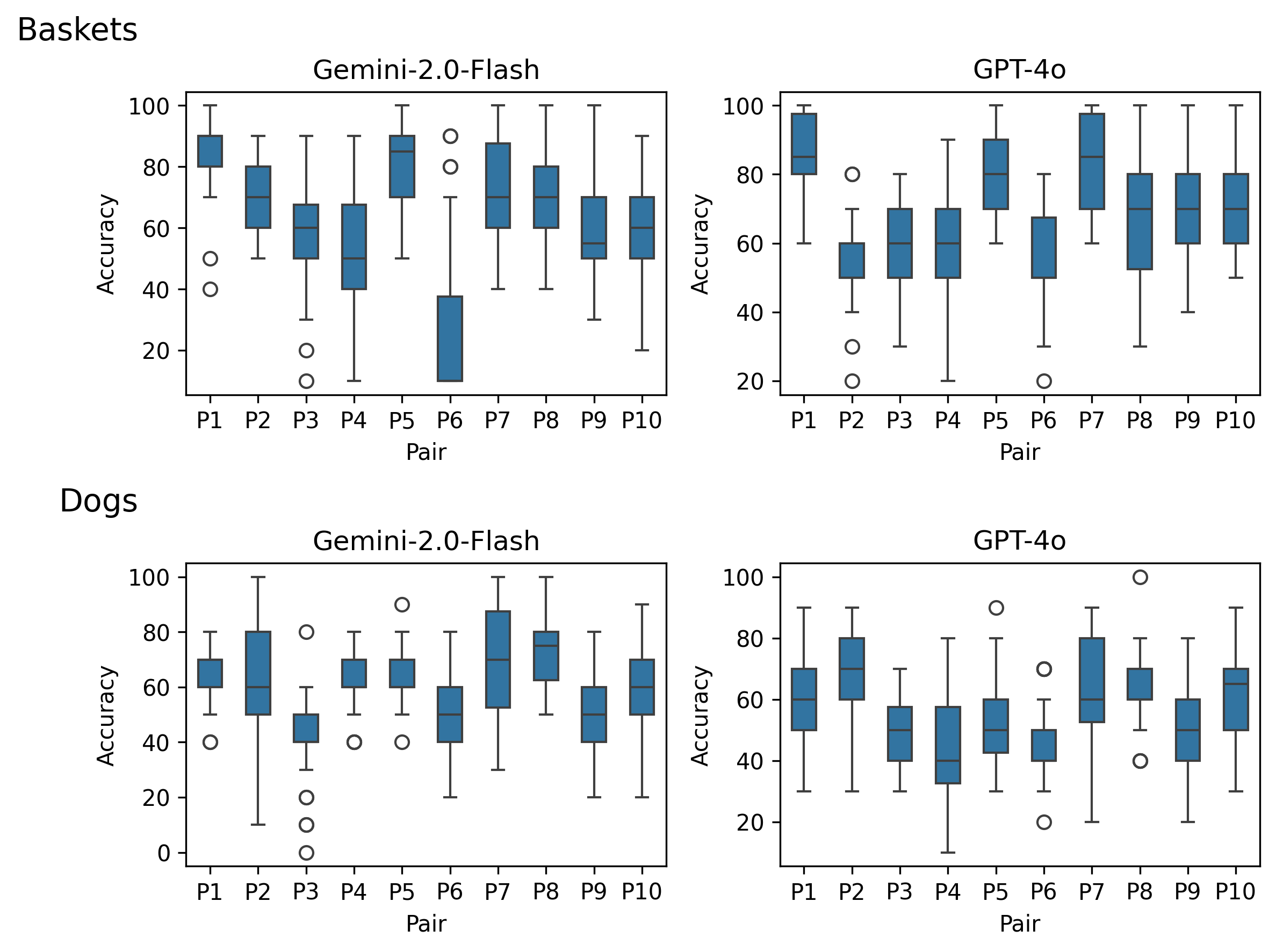}
    \caption{Accuracy boxplots of two best-performing LVLMs in the overhearer task for Round 1 conversations across 10 human pairs (whiskers denote 25th and 75th percentiles). Each boxplot represents 30 runs of a model, each with a different object ordering.}
    \label{fig:R1_30_runs_accu_boxplot}
\end{figure}
% \susan{presuming that the whiskers represent the 25th and 75th percentiles, I've added that to the figure caption --- ok Jack?} \jack{ok}.

\subsection{Robustness Analysis\label{sec:robustness}} 

% \jack{This section should just include the robustness analysis and drop the error analyses}

% Very important because otherwise, there will be concern abut fairness and trustworthiness. \jack{To rewrite.}  

% \paragraph{Robustness Analysis}  

An AI 
%good and useful 
agent should understand not only the dynamically-evolving entrained-upon expression used by a human pair for a referent, but also that
%how 
different pairs of human discourse participants entrain upon different expressions for the same referent. The agent should also be robust to object ordering in the input image. We tested matching performance on R1 for the two best-performing LVLMs, GPT-4o and Gemini-2.0-Flash, across human pairs, using different object orderings in each of 30 runs
% \Susan{ok Jack?}
for each pair. 
% \jack{This paragraph is all about reporting the robustness issues of LVLMs. We will not explain why they have these issues. We will leave that for future studies.} 

Figure~\ref{fig:R1_30_runs_accu_boxplot} shows substantial performance variation of model performance 
% \susan{as order is not the only source of variance here --- in fact, variance from different pairs of humans is probably greater --- I've added this bit. OK Jack?} 
with dialogue from different pairs of people and different object orderings.
%\jack{Owen/Susan, better now?} 
We use the difference between the 25th and 75th percentile of each boxplot as a proxy to measure performance variations and find that for the baskets (dogs) datasets, there is at least a 10.0\% (10.0\%) 
%absolute %not needed, as 25th is always lower than 75th
difference in these percentiles, with an average difference of 21.5\% (18.5\%). 

Furthermore, both models perform better on some pairs' conversations than others. For example, GPT-4o performs better with human Pair 1 than Pair 2 for their R1 conversations about baskets, with a significant difference between the means of 32.0\% (\textit{p} < 0.001). See Table~\ref{tab:meanDifferencesAcrossPairsOnR1} in Appendix~\ref{app:meanDifferenceAcrossPairs} for exhaustive pairwise comparison across pairs.
% \owen{What does Table 9 tell us that Figure 4 does not tell us and that is relevant to this paper?  Looks like significance fishing for no reason?} 

We hypothesize that 
%the across-human-pair 
these performance variations may be caused by different levels of information density in transcripts of different human pairs. However, we find no significant correlation between the average model performance and five proxy features we use, 
%including the 
namely, number of words, number of sentences, number of utterances, number of director turns, and number of matcher turns. See Table~\ref{tab:correlationAnalysesAvgPerformanceVersusInfoDensity} in Appendix~\ref{app:meanDifferenceAcrossPairs} for details. We leave further analysis for future studies.

\section{Follow-Up Experiments\label{sec:follow-up}}

We perform a series of follow-up studies to analyze factors that may affect LVLM's performance on our corpus, both in terms of a single round (R1) and over successive rounds (R1-R4). Given the relatively small size of our corpus, we focus on evaluating the \textit{out-of-box} capabilities of current LVLMs, instead of performing finetuning. 
For the best-performing closed and open LVLMs, namely GPT-4o and Qwen-2.5-VL-32B, we run follow-up experiments that vary each of
%\jack{Added "each of"}\owen{good}
the factors below. Unless otherwise stated, we run each experiment from R1 through R4 and five times for each human pair, aligning with Section~\ref{sec:methodology}. 

\paragraph{Textual Inputs}

We test this factor in four experiments.  In the first two, we vary the text that the LVLM sees; in the next two we vary how the transcript is parceled out to the LVLM. 
%task
% \jack{It is actually the transcript.}
% \susan{good -- I've changed it to transcript}
%\jack{We find the current paragraph quite lengthy, making it feel quite imbalanced in this section.}

First, we remove colloquial features to make the transcripts read more like formal written text (\textbf{+Formal}). 
%Similarly, 
% \susan{replaced Similarly with Second}
Second, we replace the 
%transcripts of interaction 
% \susan{ok jack?}
text produced incrementally during interaction with summaries of object descriptions, removing any interactive features (\textbf{-Interaction}). We rewrite the transcripts using GPT-4.1 \cite{gpt-4.1} for the two conditions and run the two LVLMs on the rewritten transcripts from R1 to R4. We inspected the output and found the generated quality acceptable for our intended use.  The prompt templates for \textbf{+Formal} and \textbf{-Interaction} can be found in Appendix~\ref{app:plus-formal} and Appendix~\ref{app:minus-interaction}, respectively.

% Second, we replace the transcripts of interaction with summaries of object descriptions, removing any interactive features (\textbf{-Interaction}) by prompting GPT-4.1 \cite{gpt-4.1} with the prompt shown in Appendix~\ref{app:minus-interaction}.  We provide the LVLM with all ten generated object descriptions at once, and again run the two LVLMs on the rewritten transcripts from R1 to R4.

Third, instead of providing the entire transcript, we provide each LVLM with complete object descriptions one at a time, manually extracted as described in Section~\ref{sec:corpus} (\textbf{ObjectDesc}). In contrast with the condition \textbf{-Interaction}, we maintain the interactive dialogue, but the LVLM only sees one object description at a time.  This provides a
control experiment with isolated descriptions, which helps disentangle whether the model performance failures are due to the overhearer setting or a more fundamental weakness in the models' basic grounding ability.
%\jack{Added this sentence}\owen{Yes, basically good} 
Since 10 times more API calls are needed for this experiment, we run it only on R1. 

Fourth, to test whether LVLMs can benefit from the ``foresight'' of accessing all transcripts at once (\textbf{AllTranscripts}), we prompt LVLMs to do the matching task for all R1-R4 transcripts at once, along with the corresponding 4 input images appended after each transcript.
%, to ensure that only one factor is varied. 

% to examine if the colloquial features of the corpus impact model performance.  

\paragraph{Visual Inputs} In the main experiments, LVLMs see 13 objects each time,
%and these objects are shuffled each time. 
in shuffled orders. In the follow-up experiments, the LVLMs see only the 10 target objects in same shuffled order
%\owen{Jack: Fixed order or random order?} \jack{We have said upfront that we vary one factor at a time, so shuffled order, but the same shuffled orders as in the main experiments.}
(\textbf{OnlyTargets}), or the 13 objects with fixed order  (\textbf{FixedOrder}). This is to determine whether either condition makes the task easier. 

\paragraph{Feedback} After an LVLM produces its answer at the end of each matching round, we provide it with the correct answers for all 10 objects in the set and prompt it to self reflect if its answers are not correct (\textbf{+Feedback}). This tests whether LVLMs can learn from feedback and improve over time.

\paragraph{Observations \& Findings} Table~\ref{tab:follow-up-Experiments} 
% \susan{I've revised Table 3's caption for clarity -- ok Jack?}
shows the performance change under each condition relative to the baseline model performance from Section~\ref{sec:results}. Removing informal and interactive features of spontaneous conversations does not yield a significant performance difference for GPT-4o and Qwen-2.5-VL-32B, but including all rounds %surprisingly,\owen{Why is this surprising?  You give it more tasks all at once.  This is the only experiment in the entire paper with multiple images, is that correct?} 
significantly \textit{hurts} model performance from Round 2 on. This means that these two models cannot effectively use information from all transcripts to resolve references across different rounds and thus do not benefit from the ``foresight'' of seeing all transcripts upfront.
% earlier conversations, which tends to be more elaborate, to resolve references in later ones, which are more efficient and reduced.
This is also true when feedback is provided to help the model reflect on mistakes, which makes no significant difference. The last two points potentially explain the lack of a consistent performance improvement over rounds observed in Section~\ref{sec:results}. That is, they show that LVLMs do not accumulate knowledge across rounds, even in light of all information presented or feedback that reveals true answers.

Moreover, both models tend to show a significant and large performance gain when given object descriptions. This shows that identifying individual objects may not be the bottleneck that causes LVLMs to perform much worse than human matchers and fail to improve over rounds.

Finally, our follow-up experiments with visual inputs demonstrate that repeating the same input image for multiple rounds typically yields no significant difference, but, as expected, models do consistently benefit from doing the matching task with only the target objects (since they can choose from 10 objects rather than 13).  
\begin{table}[]
    \centering
    \tiny

\begin{tabular}{llllll}
\toprule
 & Model & \multicolumn{2}{c}{GPT-4o} & \multicolumn{2}{c}{Qwen-2.5-VL-32B} \\
 & Source & Baskets & Dogs & Baskets & Dogs \\
Condition & Round &  &  &  &  \\
\midrule

% Utterance & 1 & -2.2 & +6.2 & {\textcolor{red}{-9.0**}} & {\textcolor{red}{-22.1***}} \\
% \noalign{\vskip 2pt} \cline{1-6} \noalign{\vskip 4pt}

\multirow[t]{4}{*}{+Formal} & 1 & -4.0 & +3.6 & -0.2 & -2.8 \\
 & 2 & +0.8 & -2.4 & -3.2 & +1.4 \\
 & 3 & -3.0 & -1.2 & -2.8 & -0.2 \\
 & 4 & -2.6 & 0.0 & -0.2 & +0.4 \\
\noalign{\vskip 2pt} \cline{1-6} \noalign{\vskip 4pt}
\multirow[t]{4}{*}{-Interaction} & 1 & +4.2 & +2.8 & +5.2 & {\textcolor{ForestGreen}{+5.8**}} \\
 & 2 & {\textcolor{ForestGreen}{+4.8*}} & -6.2 & -0.8 & -4.8 \\
 & 3 & +0.6 & -2.4 & -2.4 & -2.8 \\
 & 4 & +1.6 & -1.0 & -1.8 & -6.0 \\
\noalign{\vskip 2pt} \cline{1-6} \noalign{\vskip 4pt}

ObjectDesc & 1 & +4.4 & {\textcolor{ForestGreen}{+15.8***}} & {\textcolor{ForestGreen}{+17.0***}} & {\textcolor{ForestGreen}{+13.4***}} \\

\noalign{\vskip 2pt} \cline{1-6} \noalign{\vskip 4pt}

\multirow[t]{4}{*}{AllTranscripts} & 1 & -0.8 & -0.6 & -2.2 & -2.0 \\
 & 2 & {\textcolor{red}{-15.4***}} & {\textcolor{red}{-16.4***}} & {\textcolor{red}{-6.6**}} & {\textcolor{red}{-11.8***}} \\
 & 3 & {\textcolor{red}{-19.8***}} & {\textcolor{red}{-14.0***}} & {\textcolor{red}{-9.2***}} & {\textcolor{red}{-17.2***}} \\
 & 4 & {\textcolor{red}{-25.8***}} & {\textcolor{red}{-19.8***}} & {\textcolor{red}{-15.2***}} & {\textcolor{red}{-11.2**}} \\
\noalign{\vskip 2pt} \cline{1-6} \noalign{\vskip 4pt}
\multirow[t]{4}{*}{OnlyTargets} & 1 & {\textcolor{ForestGreen}{+11.2**}} & {\textcolor{ForestGreen}{+8.8*}} & {\textcolor{ForestGreen}{+7.6**}} & -1.2 \\
 & 2 & {\textcolor{ForestGreen}{+10.4**}} & -1.0 & {\textcolor{ForestGreen}{+7.0**}} & +5.4 \\
 & 3 & +5.2 & {\textcolor{ForestGreen}{+13.8***}} & +2.4 & +4.0 \\
 & 4 & {\textcolor{ForestGreen}{+13.6***}} & {\textcolor{ForestGreen}{+12.8***}} & {\textcolor{ForestGreen}{+12.4**}} & {\textcolor{ForestGreen}{+8.4*}} \\
\noalign{\vskip 2pt} \cline{1-6} \noalign{\vskip 4pt}
\multirow[t]{4}{*}{FixedOrder} & 1 & -3.0 & +0.8 & -0.8 & -1.4 \\
 & 2 & -0.6 & -1.6 & {\textcolor{red}{-6.8**}} & +2.4 \\
 & 3 & {\textcolor{red}{-5.0*}} & +1.8 & -4.8 & -4.0 \\
 & 4 & -2.2 & -1.4 & +1.6 & 0.0 \\
\noalign{\vskip 2pt} \cline{1-6} \noalign{\vskip 4pt}
\multirow[t]{4}{*}{+Feedback} & 1 & +3.2 & +1.6 & +1.0 & -0.2 \\
 & 2 & +1.8 & +3.8 & -0.2 & 0.0 \\
 & 3 & -0.4 & -1.0 & +0.4 & -3.4 \\
 & 4 & -1.2 & -1.8 & -1.4 & -2.0 \\
\bottomrule
\end{tabular}

    \caption{Results for the seven follow-up experiments in Section~\ref{sec:follow-up}, each differing from the main experiments (baselines) 
    %from 
    in Section~\ref{sec:results} by one factor. We highlight 
    %both 
    significant findings for both performance \textcolor{ForestGreen}{increase} and \textcolor{red}{decrease}, relative to the baseline performance (see Figure~\ref{fig:main_results}) based on paired t-tests for each condition. See Section~\ref{sec:follow-up} for details of each condition.}
    \label{tab:follow-up-Experiments}
\end{table}

\section{Conclusion}

%Despite recent advances in multimodal learning, o
Our findings demonstrate that modern LVLMs still struggle to resolve referring expressions to real-world objects produced during spontaneous conversation, a task that humans excel at when they can ground meanings together. Overhearers, whether human or LVLM, perform more poorly in a matching task than human addressees, even when they 
%hear 
are present to every word of a conversation. LVLMs in the overhearer role, even state-of-the-art models, fail to exploit the dynamic nature of conversation and do not improve over repeated referring, unlike human overhearers. These limitations constrain the practical utility of LVLMs as embodied agents, while also highlighting clear directions for future improvement. Given that our primary goal is to benchmark current LVLM capabilities in this novel overhearing setting, providing mechanistic insights or finding pathways to solutions is beyond the scope of our paper; that, we leave to future studies. We release our corpus for reproducibility and to support continued research in this area.

%Our findings demonstrate that modern LVLMs still struggle to resolve references to real-world objects produced during spontaneous conversation. In particularm speakers accrue common ground that allows them to refer to the same objects using more efficient referring expressions over time. Human overhearers understand this dynamic aspect of language use, but even state-of-the-art LVLMs fail to fully it. These limitations constrain the practical utility of LVLMs as embodied agents, while also highlighting clear directions for future improvement. We release our corpus to support continued progress in this area.
% \jack{We can add a footnote in page 1 saying that we will release the corpus upon acceptance. We can't any link it anywhere because of anonymity policy.}

\section*{Acknowledgments} 

% \jack{All: please add acknowledgments where you see fit. And how to acknowledge the seed grant? Please add.}

This material is based upon work supported by the National Science Foundation under Grant No. 2125295 and by a seed grant from Stony Brook University. Any opinions, findings, and conclusions or recommendations expressed in this material are those of the author(s) and do not necessarily reflect the views of the National Science Foundation.

Zhengxiang Wang and Owen Rambow were supported in part by funding from the Defense Advanced Research
Projects Agency (DARPA) under Contracts No.HR01121C0186, No. HR001120C0037, and PR No. HR0011154158. Any opinions, findings and conclusions or recommendations expressed in this
material are those of the authors and do not necessarily reflect the views of DARPA.

Zhengxiang Wang and Owen Rambow are grateful for support from the Institute for Advanced Computational Science (IACS) at Stony Brook University, in particular the free GPT access it provides. Zhengxiang Wang was supported by IACS's Junior Researcher Award for the academic year of 2025.

The work benefited greatly from discussions with Gregory Zelinsky, Dimitris Samaras, Ritik Raina, and Amie Paige. We also thank the five anonymous reviewers for their valuable feedback.

\section*{Limitations}

Due to resource constraints, our corpus is relatively small (80 dialogues from 10 pairs of human conversation partners). However, our corpus represents high-quality human-human spontaneous conversational data collected in a controlled experiment with a meaningful range of referential complexity (dogs and baskets) and without data contamination.  

For the same reason, we were unable to include human baselines by conducting experiments with participants acting as overhearers for the same tasks. While we agree that this would have been ideal, we make no direct statistical comparisons between LVLMs and humans as overhearers in the paper. We also note that the observations that human overhearers improve in efficiency over time, and yet perform more poorly than actual participants in the dialogue, is well established in existing psycholinguistic literature \cite{Schober1989,fox1999listening,tree2008overhearing,wilkes1992coordinating}. 

Additionally, the non-deterministic nature of large vision-language models (LVLMs) introduces challenges for evaluation; while we ran each experiment five times to reduce variability, further repetitions might yield different results. 

Our study also does not exhaustively cover all existing LVLMs—particularly some open-weight models that may perform well on these tasks—but we believe the models we selected are representative of the current state-of-the-art.

\section*{Ethical Considerations} 

\paragraph{Human Subjects} The corpus described here was collected with approval from
Stony Brook University's Institutional Review Board, CORIHS (Committe on Research Involving Human Subjects) 
and with informed consent provided by the participants. The corpus contains no personally identifiable information.

% \susan{I suggest dropping the Privacy and Risk paragraphs, as these are unrelated to the paper}
% \susan{Jack, thanks for dropping the privacy and risk paragraphs (I didn't write them).}
% \paragraph{Privacy} AI overhearing may raise a privacy concern. However, this is just an evaluation paper. Privacy is a thing between an AI agent and its user.

% \paragraph{Risk} As AI knows a specific human better with the user's consent, there may be potential risk. Either the AI may be evil, or some malicious third-party may try to hack the AI system to get private information about the user for malicious use. Again, this is an evaluation paper. 

% \Susan{can someone format and insert these} \jack{DONE}
% Clark, H. H., and Marshall, C. R. 1981. Definite reference and mutual knowledge. In Elements of discourse understanding, edited by A. K. Joshi, B. Webber, and I. Sag, 10–63. Cambridge: Cambridge University Press. 

% Reddy, M. I. (1979). The Conduit metaphor—a case of frame conflict in our language about language. In A. Ortony (Ed.), Metaphor and thought. Cambridge: Cambridge University Press.

\bibliography{custom}

\appendix

\begin{figure*}
    \centering
    \includegraphics[width=\linewidth]{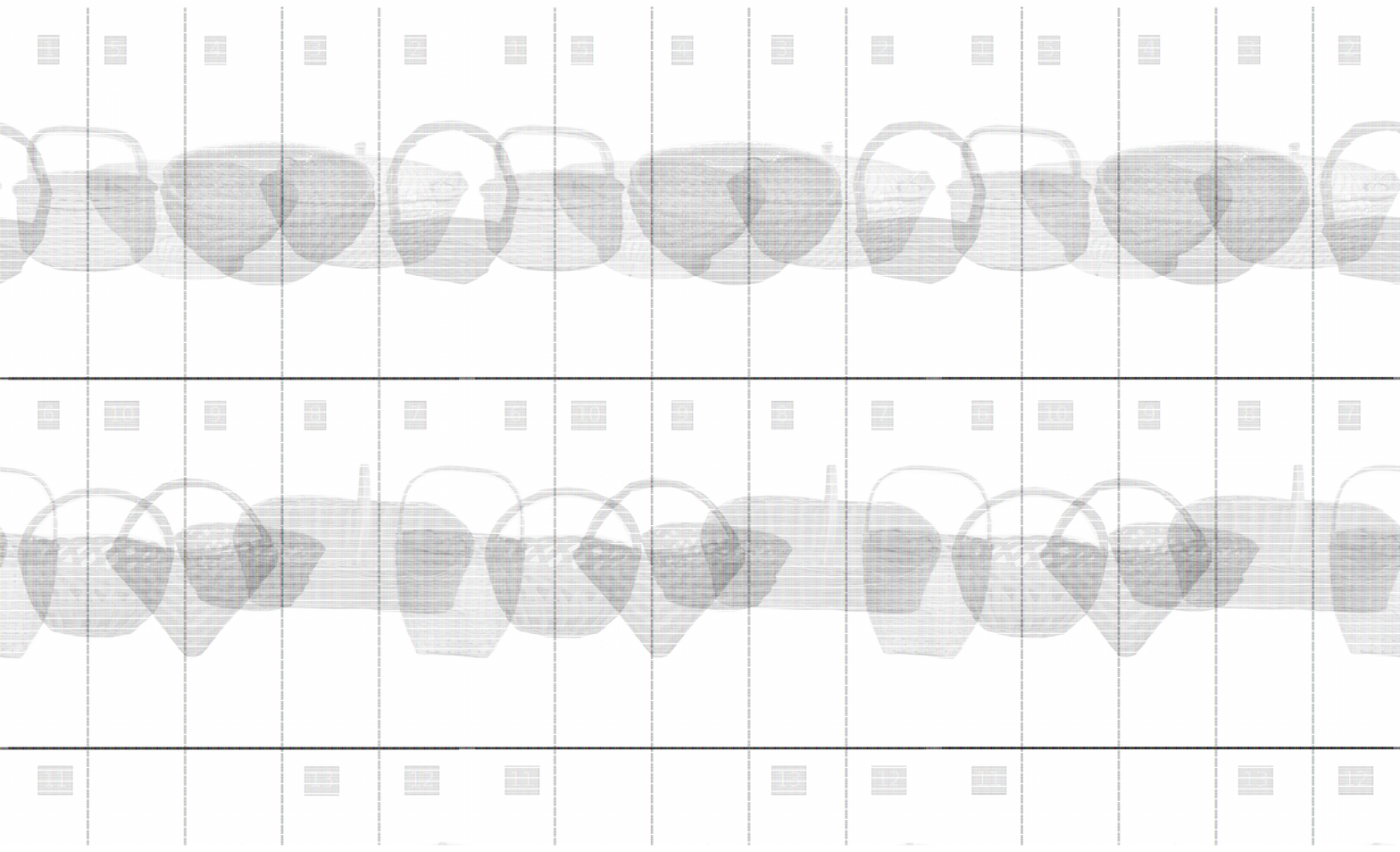}
    \caption{The 13 basket pictures from our corpus and an example input image for our experiments. The 10 target baskets are placed in the first two rows, numbered from 1 to 10, for illustration. Speakers in the original task did not see the numbers.}
    \label{fig:baskets}
\end{figure*}

\begin{figure*}
    \centering
    \includegraphics[width=\linewidth]{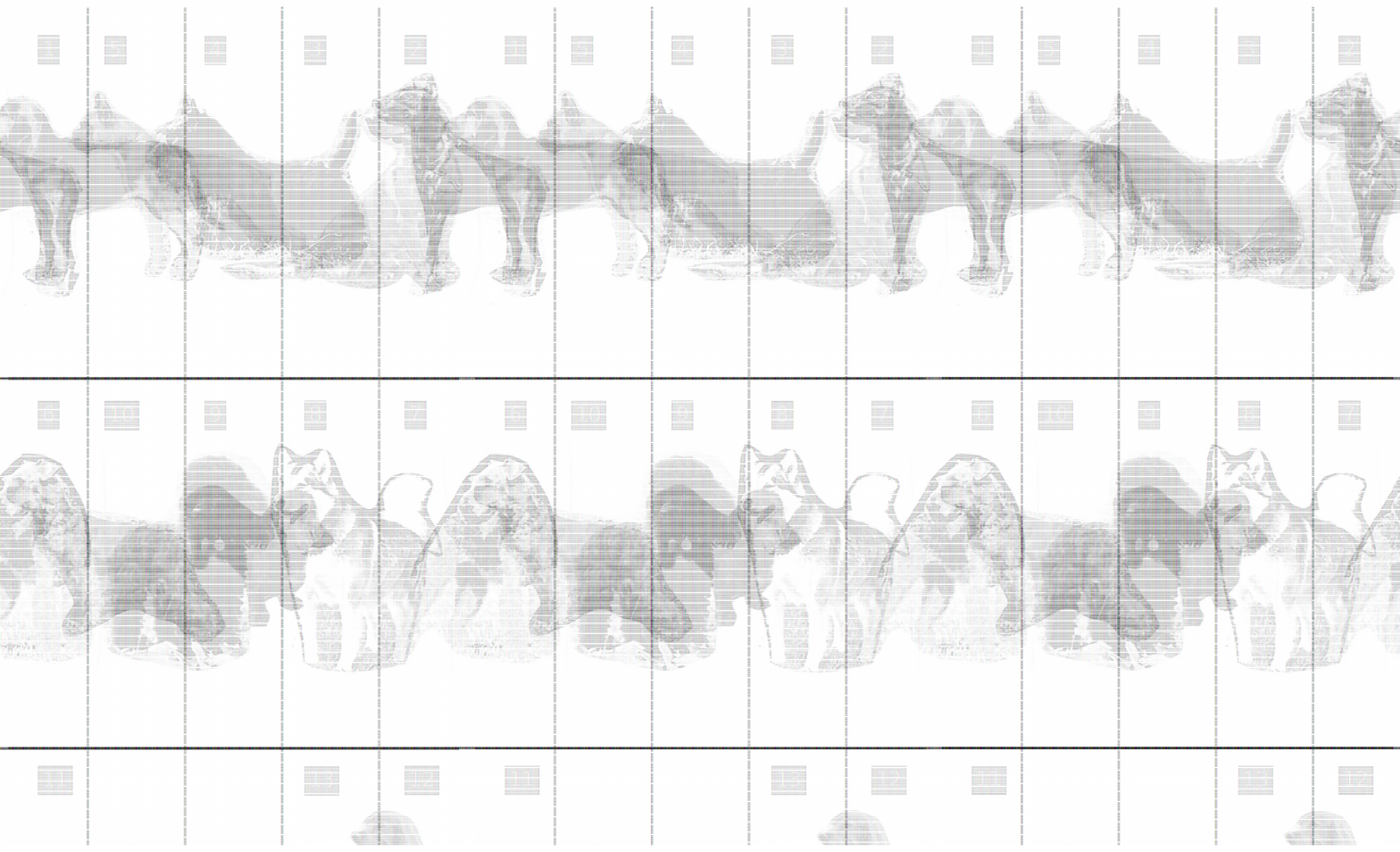}
    \caption{The 13 dog pictures from our corpus and an example input image for our experiments. The 10 target dogs are placed in the first two rows, numbered from 1 to 10, for illustration. Speakers in the original task did not see the numbers.}
    \label{fig:dogs}
\end{figure*}

\section{Corpus\label{app:corpus}}
% \susan{I've edited this section lightly}

In addition to the 80 human-to-human transcribed dialogues, our corpus also comes with 13 pictures of baskets (Figure~\ref{fig:baskets}) and 13 pictures of dogs (Figure~\ref{fig:dogs}) 
%portions of our corpus, used 
that were discussed during the original human experiments and deployed in the LVLM overhearer experiments in this paper.

%Figure \ref{fig:word count changes} shows how the average number of words used by participant pairs changes both across and within rounds for different baskets and dogs. Across rounds, descriptions became more concise, with a particularly sharp reduction after Round 1. Within-round patterns differed by object type, particularly in the most informative Round 1: for dogs, later items in the sequence tended to receive shorter descriptions, whereas for baskets, word counts showed more variability across position and sometimes increased for later items.

\begin{figure*}
    \centering
    \includegraphics[width=\linewidth]{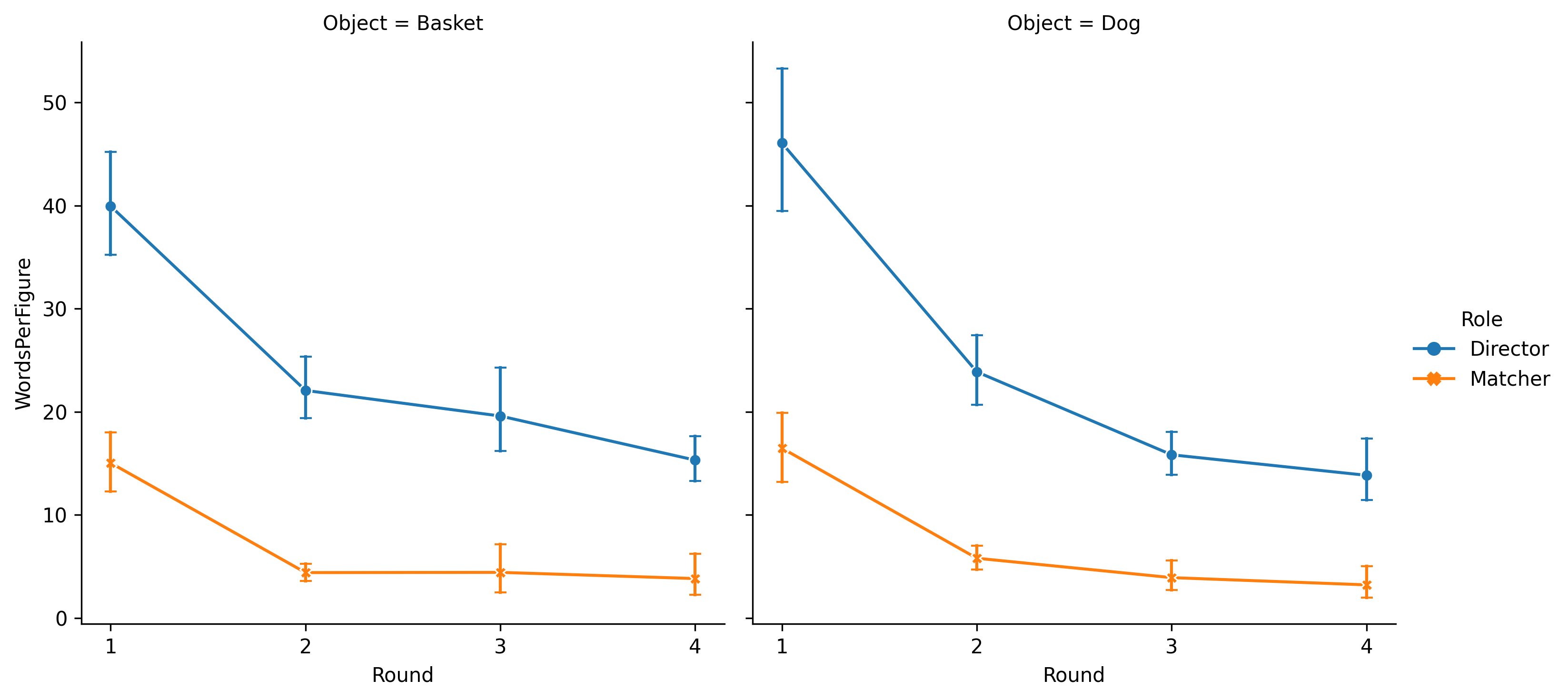}
    \caption{Mean number of words used per object across rounds for both Directors and Matchers. The plots for basket images (left) and dog images (right) both show a clear decrease in word count over time, demonstrating improved communicative efficiency. Error bars indicate $\pm$1 standard error of the mean.}
    \label{fig:word count changes per role}
\end{figure*}

To quantify how referring becomes more efficient over time,
%investigate whether communication became more efficient over time, 
we analyzed the average number of words used by directors and matchers across the four rounds for each object category (see Figure \ref{fig:word count changes per role}). Linear trend analyses revealed a significant decrease in word count as the rounds progressed. For basket images, directors' word counts significantly decreased from 39.94 words in the first round to 15.31 words in the final round, \textit{F}(1, 38) = 36.05, \textit{p} < .001. Matchers' word counts exhibited a similar pattern, decreasing from 15.04 to 3.83 words, \textit{F}(1, 38) = 19.21, \textit{p} < .001. Comparable trends were observed for dog images. Directors' word counts dropped from 46.05 to 13.84 words, \textit{F}(1, 38) = 37.96, \textit{p} < .001, while matchers' word counts decreased from 16.47 to 3.22 words, \textit{F}(1, 38) = 39.90, \textit{p} < .001.

We also examined whether the object category influenced word counts (a proxy for effort) by comparing the average words used per round for baskets and dogs. On average, pairs used 31.16 words per round to describe baskets and 32.25 for dogs. A paired sample t-test found this difference was not statistically significant, \textit{t}(9) = -0.34, \textit{p} = 0.74.

\section{Experiments\label{app:experiments}}

\subsection{Details about LVLMs \label{app:models}}

The specific versions of LVLMs used in our study are as follows. We used the APIs from the respective LVLM providers for the first four proprietary LVLMs. The open-weight models were deployed locally through Hugging Face and run on 3 Nvidia RTX A6000s.

\begin{itemize}
    \item GPT-4o: \textsc{gpt-4o-2024-08-06}
    \item GPT-4o-mini: \textsc{gpt-4o-mini-2024-07-18}
    \item Gemini-2.0-Flash: Accessed in April 2025.
    \item Claude-3.7-Sonnet: \textsc{claude-3-7-sonnet-20250219}.
    \item Qwen-2.5-VL-7B: Hugging Face model card, \textsc{Qwen/Qwen2.5-VL-7B-Instruct}
    \item Qwen-2.5-VL-32B: Hugging Face model card, \textsc{Qwen/Qwen2.5-VL-32B-Instruct}
    \item Pixtral-12B: Hugging Face model card, \textsc{mistralai/Pixtral-12B-2409}
\end{itemize}

\subsection{Input Images}

We generated 30 different orderings of objects for both baskets and dogs datasets, in addition to the two orderings shown in Figure~\ref{fig:baskets} and Figure~\ref{fig:dogs}, respectively. These additional images share the same layout as the ones in Figures~\ref{fig:baskets} and~\ref{fig:dogs} and differ from the latter only in the specific order of the 13 objects.

To ensure that all LVLMs see the same input images across multiple rounds as well as across different human pairs to allow for paired t-tests, we create a playbook that stores the the specific input image for each run number and for each round of conversation (1-4). In the main experiments, we run each model on each transcript from each human pair for five times and the model see exactly the same input image for each round number across 10 human pairs for both baskets and dogs datasets. The playbook was simply created by randomly sampling four distinct images from the generated images for 100 times, although we only ran each model up to 30 times in our study (see Section~\ref{sec:robustness}).
%\susan{This paragraph isn't entirely clear to me yet; we can clarify it in the final version.}

\begin{table}[]
    \centering
    \scriptsize
    
\begin{tabular}{lllll}
\toprule
 & Starting Round & R1 & R2 & R3 \\
Source & Model &  &  &  \\
\midrule
\multirow[t]{7}{*}{Baskets} & Claude-3.7-Sonnet & {\textcolor{ForestGreen}{0.2*}} & {\textcolor{ForestGreen}{0.3***}} & {\textcolor{ForestGreen}{0.3**}} \\
 & Gemini-2.0-Flash & {\textcolor{ForestGreen}{0.2*}} & 0.1 & 0.1 \\
 & GPT-4o & {\textcolor{ForestGreen}{0.1*}} & 0.1 & 0.1 \\
 & GPT-4o-mini & {\textcolor{ForestGreen}{0.2**}} & {\textcolor{ForestGreen}{0.2*}} & {\textcolor{ForestGreen}{0.4***}} \\
 & Qwen-2.5-VL-32B & 0.1 & -0.1 & 0.0 \\
 & Qwen-2.5-VL-7B & {\textcolor{red}{-0.3***}} & {\textcolor{red}{-0.4***}} & -0.1 \\
 & Pixtral-12B & 0.1 & {\textcolor{red}{-0.3***}} & {\textcolor{ForestGreen}{0.3**}} \\

\noalign{\vskip 2pt} \cline{1-5} \noalign{\vskip 4pt}

\multirow[t]{7}{*}{Dogs} & Claude-3.7-Sonnet & {\textcolor{ForestGreen}{0.2*}} & 0.1 & 0.1 \\
 & Gemini-2.0-Flash & {\textcolor{ForestGreen}{0.2**}} & 0.1 & {\textcolor{ForestGreen}{0.3**}} \\
 & GPT-4o & {\textcolor{ForestGreen}{0.2**}} & 0.0 & 0.2 \\
 & GPT-4o-mini & 0.0 & -0.0 & -0.0 \\
 & Qwen-2.5-VL-32B & 0.1 & 0.0 & -0.0 \\
 & Qwen-2.5-VL-7B & 0.0 & {\textcolor{ForestGreen}{0.2*}} & -0.2 \\
 & Pixtral-12B & -0.0 & -0.0 & -0.1 \\

\bottomrule
\end{tabular}

    \caption{Overall performance trend for each LVLM starting at R$_i$, using Spearman rank correlation. We highlight both \textbf{significant} \textcolor{ForestGreen}{positive} and \textcolor{red}{negative} trends and use asterisks to denote different levels of significance, where ``*'' means \textit{p} < 0.05, ``**'' means \textit{p} < 0.01, and ``**'' means \textit{p} < 0.001.}
    \label{tab:OverallPerformanceTrendsSpearman}
\end{table}

\begin{table}[]
    \centering
    \scriptsize
    
\begin{tabular}{lllll}
\toprule
 & Starting Round & R1 & R2 & R3 \\
Source & Model &  &  &  \\
\midrule

\multirow[t]{7}{*}{Baskets} & Claude-3.7-Sonnet & {\textcolor{ForestGreen}{0.1*}} & {\textcolor{ForestGreen}{0.3***}} & {\textcolor{ForestGreen}{0.3**}} \\
 & Gemini-2.0-Flash & {\textcolor{ForestGreen}{0.1*}} & 0.1 & 0.1 \\
 & GPT-4o & {\textcolor{ForestGreen}{0.1*}} & 0.1 & 0.1 \\
 & GPT-4o-mini & {\textcolor{ForestGreen}{0.2**}} & {\textcolor{ForestGreen}{0.2*}} & {\textcolor{ForestGreen}{0.4***}} \\
 & Qwen-2.5-VL-32B & 0.1 & -0.1 & 0.0 \\
 & Qwen-2.5-VL-7B & {\textcolor{red}{-0.2***}} & {\textcolor{red}{-0.4***}} & -0.1 \\
 & Pixtral-12B & 0.1 & {\textcolor{red}{-0.2***}} & {\textcolor{ForestGreen}{0.3**}} \\

\noalign{\vskip 2pt} \cline{1-5} \noalign{\vskip 4pt}

\multirow[t]{7}{*}{Dogs} & Claude-3.7-Sonnet & {\textcolor{ForestGreen}{0.1*}} & 0.1 & 0.1 \\
 & Gemini-2.0-Flash & {\textcolor{ForestGreen}{0.2**}} & 0.1 & {\textcolor{ForestGreen}{0.2**}} \\
 & GPT-4o & {\textcolor{ForestGreen}{0.2**}} & 0.0 & 0.1 \\
 & GPT-4o-mini & 0.0 & -0.0 & -0.0 \\
 & Qwen-2.5-VL-32B & 0.1 & 0.0 & -0.0 \\
 & Qwen-2.5-VL-7B & 0.0 & {\textcolor{ForestGreen}{0.2*}} & -0.2 \\
 & Pixtral-12B & -0.0 & -0.0 & -0.1 \\

\bottomrule
\end{tabular}

    \caption{Overall performance trend for each LVLM starting at R$_i$, using Kendall rank correlation. We highlight both \textbf{significant} \textcolor{ForestGreen}{positive} and \textcolor{red}{negative} trends, using ``*'' to denote different levels of significance, where ``*'' means \textit{p} < 0.05, ``**'' means \textit{p} < 0.01, and ``**'' means \textit{p} < 0.001.}
    \label{tab:OverallPerformanceTrendsKendall}
\end{table}

\section{Results\label{app:results}}

\begin{table*}[]
    \centering

\begin{tabular}{lllll}
\toprule
 & Starting Round & R1 & R2 & R3 \\
Source & Model &  &  &  \\
\midrule
\multirow[t]{7}{*}{Baskets} & Claude-3.7-Sonnet & 36.0 / 24.0 / 4.0 & 56.0 / 50.0 / 12.0 & 86.0 / 58.0 \\
 & Gemini-2.0-Flash & 22.0 / 22.0 / 2.0 & 34.0 / 32.0 / 6.0 & 76.0 / 44.0 \\
 & GPT-4o & 26.0 / 26.0 / 6.0 & 42.0 / 32.0 / 2.0 & 72.0 / 46.0 \\
 & GPT-4o-mini & 18.0 / 16.0 / 0.0 & 36.0 / 32.0 / 6.0 & 84.0 / 66.0 \\
 & Qwen-2.5-VL-32B & 10.0 / 10.0 / 2.0 & 26.0 / 16.0 / 6.0 & 62.0 / 40.0 \\
 & Qwen-2.5-VL-7B & 6.0 / 4.0 / 0.0 & 14.0 / 2.0 / 0.0 & 62.0 / 22.0 \\
 & Pixtral-12B & 18.0 / 16.0 / 2.0 & 22.0 / 8.0 / 0.0 & 82.0 / 44.0 \\

\midrule \midrule

\multirow[t]{7}{*}{Dogs} & Claude-3.7-Sonnet & 46.0 / 30.0 / 2.0 & 54.0 / 30.0 / 4.0 & 84.0 / 48.0 \\
 & Gemini-2.0-Flash & 36.0 / 32.0 / 2.0 & 56.0 / 38.0 / 6.0 & 88.0 / 60.0 \\
 & GPT-4o & 22.0 / 20.0 / 0.0 & 52.0 / 36.0 / 8.0 & 85.4 / 52.1 \\
 & GPT-4o-mini & 12.0 / 12.0 / 4.0 & 26.0 / 26.0 / 6.0 & 60.4 / 31.2 \\
 & Qwen-2.5-VL-32B & 14.0 / 14.0 / 2.0 & 22.0 / 20.0 / 8.0 & 60.0 / 40.0 \\
 & Qwen-2.5-VL-7B & 10.0 / 10.0 / 0.0 & 44.0 / 34.0 / 18.0 & 54.0 / 22.0 \\
 & Pixtral-12B & 28.0 / 28.0 / 0.0 & 42.0 / 30.0 / 2.0 & 64.0 / 36.0 \\

\bottomrule
\end{tabular}
    
    \caption{Percentage of time an LVLM's performance monotonically increases from a starting round (R1-R3) to the end round (R4). We report the following numbers in the table, separated by ``/'' in each cell: percentage of of monotonically increasing, percentage of of monotonically increasing with a positive slope (the model performance on the end round must be greater than the starting round), and percentage of strictly monotonically increasing (model performance on each round is strictly greater than that on the previous round). For a R3 start, there are only 2 rounds (R3 and R4), so the last two numbers are identical and we only report one in the R3 column. Again, proprietary LVLMs are overall more likely to show monotonically increasing performance.}
    \label{tab:performanceMonotonicity}
\end{table*}

\begin{table}[]
    \centering
    \small 

\begin{tabular}{l|ccc|ccc}
\toprule
Source & \multicolumn{3}{c}{\textbf{Baskets}} & \multicolumn{3}{c}{\textbf{Dogs}} \\
 & R1 & R2 & R3 & R1 & R2 & R3 \\
\midrule
Claude-3.7-Sonnet & 20 & 30 & 20 & 10 & 0 & 0 \\
Gemini-2.0-Flash & 30 & 10 & 0 & 10 & 0 & 10 \\
GPT-4o & 10 & 0 & 0 & 10 & 0 & 0 \\
GPT-4o-mini & 30 & 20 & 20 & 0 & 0 & 0 \\
Qwen-2.5-VL-32B & 0 & 0 & 0 & 20 & 0 & 10 \\
Qwen-2.5-VL-7B & 0 & 0 & 0 & 0 & 20 & 0 \\
Pixtral-12B & 0 & 0 & 10 & 0 & 0 & 0 \\
\bottomrule
\end{tabular}
    
    \caption{Percentage of human pairs for which LVLMs show a consistent improvement when starting at R$_{i}$.}
    \label{tab:performanceOverTimeAcrossHumanPairs}
\end{table}

\subsection{Overall Performance Trend\label{app:overallPerformanceTrend}}

Table~\ref{tab:OverallPerformanceTrendsSpearman} and Table~\ref{tab:OverallPerformanceTrendsKendall} shows the overall performance trend for each LVLM across baskets and dogs using Spearman rank correlation and Kendall rank correlation, respectively. The overall results are consistent with Table~\ref{tab:performanceOverRounds} in Section~\ref{sec:results}.

Table~\ref{tab:performanceOverTimeAcrossHumanPairs} shows the percentage of human pairs for whom LVLMs show a consistent improvement when starting at R$_i$. We use OLS regression to measure if there is a consistent improvement (i.e., significant positive coefficient).

Table~\ref{tab:performanceMonotonicity} shows the percentage of time an LVLM’s performance monotonically increases from a starting round (R1-R3) to the
end round (R4) over all runs of each model. We report three levels of ``monotonically increases'': (1) monotonically increasing or non-decreasing, (2) monotonically increasing with a positive slope, meaning that model performance at the end round must surpass the starting round, and (3) strictly monotonically increasing, where we require that model performance on a round must be better than that of a previous round. As can be seen in Table~\ref{tab:performanceMonotonicity}, if we use the strictest measurement, none of the tested LVLMs show more than 6\% strictly monotonically increasing performance.

\begin{table*}[]
    \centering
    \small

\begin{tabular}{llllll}
\toprule
 &  & Late Start  $\rightarrow$ & R2-R4 & R3-R4 & R4 \\
Source & Model & Early Start $\downarrow$ &  &  &  \\
\midrule
\multirow[t]{21}{*}{Baskets}

 & \multirow[t]{3}{*}{Claude-3.7-Sonnet} & R1-R4 & 1.4 & {\textcolor{ForestGreen}{11.6***}} & {\textcolor{ForestGreen}{15.6***}} \\
 &  & R2-R4 & - & {\textcolor{ForestGreen}{13.0***}} & {\textcolor{ForestGreen}{19.0***}} \\
 &  & R3-R4 & - & - & {\textcolor{ForestGreen}{8.0*}} \\
\noalign{\vskip 2pt} \cline{2-6} \noalign{\vskip 4pt}

 & \multirow[t]{3}{*}{Gemini-2.0-Flash} & R1-R4 & {\textcolor{ForestGreen}{3.1*}} & {\textcolor{ForestGreen}{10.2***}} & {\textcolor{ForestGreen}{10.4**}} \\
 &  & R2-R4 & - & {\textcolor{ForestGreen}{9.7***}} & {\textcolor{ForestGreen}{12.2***}} \\
 &  & R3-R4 & - & - & 2.0 \\
\noalign{\vskip 2pt} \cline{2-6} \noalign{\vskip 4pt}

& \multirow[t]{3}{*}{GPT-4o} & R1-R4 & {\textcolor{ForestGreen}{8.4***}} & {\textcolor{ForestGreen}{17.9***}} & {\textcolor{ForestGreen}{15.6***}} \\
 &  & R2-R4 & - & {\textcolor{ForestGreen}{9.2***}} & {\textcolor{ForestGreen}{9.0**}} \\
 &  & R3-R4 & - & - & -0.2 \\
\noalign{\vskip 2pt} \cline{2-6} \noalign{\vskip 4pt}

 & \multirow[t]{3}{*}{GPT-4o-mini} & R1-R4 & {\textcolor{ForestGreen}{3.9**}} & {\textcolor{ForestGreen}{4.4*}} & 4.2 \\
 &  & R2-R4 & - & 3.6 & 0.0 \\
 &  & R3-R4 & - & - & 0.8 \\
\noalign{\vskip 2pt} \cline{2-6} \noalign{\vskip 4pt}

 & \multirow[t]{3}{*}{Qwen-2.5-VL-32B} & R1-R4 & 2.7 & 0.8 & -1.4 \\
 &  & R2-R4 & - & -1.8 & -3.6 \\
 &  & R3-R4 & - & - & 1.4 \\ \noalign{\vskip 2pt} \cline{2-6} \noalign{\vskip 4pt}

 & \multirow[t]{3}{*}{Qwen-2.5-VL-7B} & R1-R4 & {\textcolor{red}{-4.0**}} & {\textcolor{red}{-2.9*}} & {\textcolor{red}{-10.4***}} \\
 &  & R2-R4 & - & -1.2 & {\textcolor{red}{-9.8***}} \\
 &  & R3-R4 & - & - & {\textcolor{red}{-8.6***}} \\ \noalign{\vskip 2pt} \cline{2-6} \noalign{\vskip 4pt}
 
 & \multirow[t]{3}{*}{Pixtral-12B} & R1-R4 & 1.9 & 2.3 & -1.2 \\
 &  & R2-R4 & - & -1.1 & {\textcolor{red}{-4.4**}} \\
 &  & R3-R4 & - & - & -1.2 \\

\midrule \midrule

\multirow[t]{21}{*}{Dogs} 

 & \multirow[t]{3}{*}{Claude-3.7-Sonnet} & R1-R4 & 2.9 & {\textcolor{ForestGreen}{9.6***}} & {\textcolor{ForestGreen}{16.4***}} \\
 &  & R2-R4 & - & {\textcolor{ForestGreen}{7.0**}} & {\textcolor{ForestGreen}{12.4***}} \\
 &  & R3-R4 & - & - & {\textcolor{ForestGreen}{8.6**}} \\
\noalign{\vskip 2pt} \cline{2-6} \noalign{\vskip 4pt}

 & \multirow[t]{3}{*}{Gemini-2.0-Flash} & R1-R4 & 2.3 & {\textcolor{ForestGreen}{15.9***}} & {\textcolor{ForestGreen}{8.0**}} \\
 &  & R2-R4 & - & {\textcolor{ForestGreen}{13.9***}} & {\textcolor{ForestGreen}{8.8*}} \\
 &  & R3-R4 & - & - & -2.4 \\
\noalign{\vskip 2pt} \cline{2-6} \noalign{\vskip 4pt}

& \multirow[t]{3}{*}{GPT-4o} & R1-R4 & 2.2 & {\textcolor{ForestGreen}{16.1***}} & {\textcolor{ForestGreen}{11.4***}} \\
 &  & R2-R4 & - & {\textcolor{ForestGreen}{15.9***}} & {\textcolor{ForestGreen}{9.6**}} \\
 &  & R3-R4 & - & - & -2.1 \\
\noalign{\vskip 2pt} \cline{2-6} \noalign{\vskip 4pt}

 & \multirow[t]{3}{*}{GPT-4o-mini} & R1-R4 & 0.3 & {\textcolor{red}{-4.4*}} & 0.0 \\
 &  & R2-R4 & - & {\textcolor{red}{-5.6**}} & -3.4 \\
 &  & R3-R4 & - & - & 2.9 \\
\noalign{\vskip 2pt} \cline{2-6} \noalign{\vskip 4pt}

 & \multirow[t]{3}{*}{Qwen-2.5-VL-32B} & R1-R4 & 0.9 & 3.8 & {\textcolor{ForestGreen}{10.2**}} \\
 &  & R2-R4 & - & 0.5 & {\textcolor{ForestGreen}{7.8*}} \\
 &  & R3-R4 & - & - & {\textcolor{ForestGreen}{8.0**}} \\

\noalign{\vskip 2pt} \cline{2-6} \noalign{\vskip 4pt}

 & \multirow[t]{3}{*}{Qwen-2.5-VL-7B} & R1-R4 & {\textcolor{red}{-12.8***}} & {\textcolor{red}{-5.2***}} & -4.0 \\
 &  & R2-R4 & - & {\textcolor{ForestGreen}{8.8***}} & {\textcolor{ForestGreen}{12.8***}} \\
 &  & R3-R4 & - & - & 1.0 \\
\noalign{\vskip 2pt} \cline{2-6} \noalign{\vskip 4pt}

 & \multirow[t]{3}{*}{Pixtral-12B} & R1-R4 & -2.3 & -1.3 & -1.4 \\
 &  & R2-R4 & - & -1.1 & -2.0 \\
 &  & R3-R4 & - & - & -0.2 \\

\bottomrule
\end{tabular}

    \caption{Pairwise mean differences of performance on overlapping rounds between an early start (R1-R4-R3-R4 in the rows) and a late start (R2-R4-R4 in the columns). For example, the overlapping rounds for R1-R4 and R2-R4 are R2-R3-R4, and for R2-R4 and R3-R4 they are R3-R4. We use paired t-tests to determine if there is a significant difference between two different starts and highlight both \textbf{significant} \textcolor{ForestGreen}{positive} and \textcolor{red}{negative} mean differences. We indicate significance level using asterisks: ``*'' means \textit{p} < 0.05, ``**'' means \textit{p} < 0.01, and ``**'' means \textit{p} < 0.001. In each round, there are 50 experiments (10 human pairs times 5 runs of an LVLM), so the degree of freedom is 100 times ``number of overlapping rounds'' minus 1.}
    \label{tab:meanDiffBetweenOverlappingRounds}
\end{table*}

\begin{table*}[]
    \centering
    \small

\begin{tabular}{llccccc}
\toprule
 &  & Late Start  $\rightarrow$ & R2 & R3 & R4 \\
Source & Model & Early Start  $\downarrow$ &  &  &  \\
\midrule
\multirow[t]{21}{*}{Baskets} 
& \multirow[t]{3}{*}{Claude-3.7-Sonnet} & R1 & 1.2 & {\textcolor{ForestGreen}{11.2***}} & {\textcolor{ForestGreen}{9.8**}} \\
 &  & R2 & - & {\textcolor{ForestGreen}{10.0***}} & {\textcolor{ForestGreen}{8.6**}} \\
 &  & R3 & - & - & -1.4 \\

\noalign{\vskip 2pt}
\cline{2-6}
\noalign{\vskip 4pt}

 & \multirow[t]{3}{*}{Gemini-2.0-Flash} & R1 & {\textcolor{red}{-7.0*}} & 2.2 & -0.4 \\
 &  & R2 & - & {\textcolor{ForestGreen}{9.2**}} & {\textcolor{ForestGreen}{6.6**}} \\
 &  & R3 & - & - & -2.6 \\
\noalign{\vskip 2pt}
\cline{2-6}
\noalign{\vskip 4pt}

& \multirow[t]{3}{*}{GPT-4o} & R1 & 4.8 & {\textcolor{ForestGreen}{14.0**}} & {\textcolor{ForestGreen}{10.2**}} \\
 &  & R2 & - & {\textcolor{ForestGreen}{9.2*}} & 5.4 \\
 &  & R3 & - & - & -3.8 \\

\noalign{\vskip 2pt}
\cline{2-6}
\noalign{\vskip 4pt}

 & \multirow[t]{3}{*}{GPT-4o-mini} & R1 & -2.2 & 4.0 & {\textcolor{red}{-8.4**}} \\
 &  & R2 & - & {\textcolor{ForestGreen}{6.2*}} & {\textcolor{red}{-6.2*}} \\
 &  & R3 & - & - & {\textcolor{red}{-12.4***}} \\

\noalign{\vskip 2pt}
\cline{2-6}
\noalign{\vskip 4pt}

 & \multirow[t]{3}{*}{Qwen-2.5-VL-32B} & R1 & {\textcolor{red}{-6.4*}} & -4.8 & -5.4 \\
 &  & R2 & - & 1.6 & 1.0 \\
 &  & R3 & - & - & -0.6 \\

\noalign{\vskip 2pt}
\cline{2-6}
\noalign{\vskip 4pt}

 & \multirow[t]{3}{*}{Qwen-2.5-VL-7B} & R1 & {\textcolor{red}{-9.6***}} & 0.8 & {\textcolor{red}{-4.8*}} \\
 &  & R2 & - & {\textcolor{ForestGreen}{10.4***}} & 4.8 \\
 &  & R3 & - & - & {\textcolor{red}{-5.6*}} \\ 
\noalign{\vskip 2pt}
\cline{2-6}
\noalign{\vskip 4pt}

 & \multirow[t]{3}{*}{Pixtral-12B} & R1 & {\textcolor{red}{-4.6*}} & 1.8 & {\textcolor{red}{-3.6*}} \\
 &  & R2 & - & {\textcolor{ForestGreen}{6.4***}} & 1.0 \\
 &  & R3 & - & - & {\textcolor{red}{-5.4**}} \\
 
\midrule \midrule

\multirow[t]{21}{*}{Dogs}  

& \multirow[t]{3}{*}{Claude-3.7-Sonnet} & R1 & -0.6 & 5.4 & {\textcolor{ForestGreen}{9.0**}} \\
 &  & R2 & - & 6.0 & {\textcolor{ForestGreen}{9.6*}} \\
 &  & R3 & - & - & 3.6 \\
\noalign{\vskip 2pt}
\cline{2-6}
\noalign{\vskip 4pt}

 & \multirow[t]{3}{*}{Gemini-2.0-Flash} & R1 & -6.4 & {\textcolor{ForestGreen}{9.6**}} & -2.8 \\
 &  & R2 & - & {\textcolor{ForestGreen}{16.0***}} & 3.6 \\
 &  & R3 & - & - & {\textcolor{red}{-12.4***}} \\
\noalign{\vskip 2pt}
\cline{2-6}
\noalign{\vskip 4pt}

& \multirow[t]{3}{*}{GPT-4o} & R1 & -6.4 & {\textcolor{ForestGreen}{9.4**}} & -0.8 \\
 &  & R2 & - & {\textcolor{ForestGreen}{15.8***}} & 5.6 \\
 &  & R3 & - & - & {\textcolor{red}{-10.2**}} \\
\noalign{\vskip 2pt}
\cline{2-6}
\noalign{\vskip 4pt}

 & \multirow[t]{3}{*}{GPT-4o-mini} & R1 & -4.4 & {\textcolor{red}{-5.9*}} & -1.6 \\
 &  & R2 & - & -1.5 & 2.8 \\
 &  & R3 & - & - & 4.3 \\
\noalign{\vskip 2pt}
\cline{2-6}
\noalign{\vskip 4pt}

 & \multirow[t]{3}{*}{Qwen-2.5-VL-32B} & R1 & -4.8 & -1.6 & 5.2 \\
 &  & R2 & - & 3.2 & {\textcolor{ForestGreen}{10.0*}} \\
 &  & R3 & - & - & {\textcolor{ForestGreen}{6.8*}} \\

\noalign{\vskip 2pt}
\cline{2-6}
\noalign{\vskip 4pt}

 & \multirow[t]{3}{*}{Qwen-2.5-VL-7B} & R1 & -10.2 & {\textcolor{red}{-8.6**}} & -2.6 \\
 &  & R2 & - & 1.6 & 7.6 \\
 &  & R3 & - & - & {\textcolor{ForestGreen}{6.0*}} \\
 
\noalign{\vskip 2pt}
\cline{2-6}
\noalign{\vskip 4pt}

 & \multirow[t]{3}{*}{Pixtral-12B} & R1 & {\textcolor{red}{-12.4*}} & 0.0 & 0.0 \\
 &  & R2 & - & {\textcolor{ForestGreen}{12.4*}} & {\textcolor{ForestGreen}{12.4*}} \\
 &  & R3 & - & - & 0.0 \\

\bottomrule
\end{tabular}

    \caption{Pairwise mean differences between an early start (R1-R3 in the rows) and a late start (R2-R4 in the columns). We use paired t-tests to determine if there is a significant difference between two different starts and highlight both \textbf{significant} \textcolor{ForestGreen}{positive} and \textcolor{red}{negative} mean differences. We indicate significance level using asterisks: ``*'' means \textit{p} < 0.05, ``**'' means \textit{p} < 0.01, and ``**'' means \textit{p} < 0.001. In each round, there are 50 experiments (10 human pairs times 5 runs of an LVLM), so the degree of freedom is 99.}
    \label{tab:startingRoundsDifsTTestsBaskets}
\end{table*}

\begin{table*}[]
    \centering
    \scriptsize

\begin{tabular}{llllllllllll}
\toprule
 &  & Pair2 $\rightarrow$ & P2 & P3 & P4 & P5 & P6 & P7 & P8 & P9 & P10 \\
Source & Model & Pair1 $\downarrow$ &  &  &  &  &  &  &  &  &  \\
\midrule
\multirow[t]{18}{*}{Baskets} & \multirow[t]{9}{*}{GPT-4o} & P1 & {\textcolor{ForestGreen}{32.0***}} & {\textcolor{ForestGreen}{29.0***}} & {\textcolor{ForestGreen}{25.0***}} & {\textcolor{ForestGreen}{6.0**}} & {\textcolor{ForestGreen}{31.7***}} & 3.7 & {\textcolor{ForestGreen}{19.7***}} & {\textcolor{ForestGreen}{17.7***}} & {\textcolor{ForestGreen}{15.7***}} \\
 &  & P2 & - & -3.0 & -7.0 & {\textcolor{red}{-26.0***}} & -0.3 & {\textcolor{red}{-28.3***}} & {\textcolor{red}{-12.3**}} & {\textcolor{red}{-14.3***}} & {\textcolor{red}{-16.3***}} \\
 &  & P3 & - & - & -4.0 & {\textcolor{red}{-23.0***}} & 2.7 & {\textcolor{red}{-25.3***}} & {\textcolor{red}{-9.3*}} & {\textcolor{red}{-11.3**}} & {\textcolor{red}{-13.3***}} \\
 &  & P4 & - & - & - & {\textcolor{red}{-19.0***}} & 6.7 & {\textcolor{red}{-21.3***}} & -5.3 & -7.3 & {\textcolor{red}{-9.3*}} \\
 &  & P5 & - & - & - & - & {\textcolor{ForestGreen}{25.7***}} & -2.3 & {\textcolor{ForestGreen}{13.7***}} & {\textcolor{ForestGreen}{11.7***}} & {\textcolor{ForestGreen}{9.7**}} \\
 &  & P6 & - & - & - & - & - & {\textcolor{red}{-28.0***}} & {\textcolor{red}{-12.0**}} & {\textcolor{red}{-14.0**}} & {\textcolor{red}{-16.0***}} \\
 &  & P7 & - & - & - & - & - & - & {\textcolor{ForestGreen}{16.0***}} & {\textcolor{ForestGreen}{14.0***}} & {\textcolor{ForestGreen}{12.0***}} \\
 &  & P8 & - & - & - & - & - & - & - & -2.0 & -4.0 \\
 &  & P9 & - & - & - & - & - & - & - & - & -2.0 \\

\noalign{\vskip 2pt} \cline{2-12} \noalign{\vskip 4pt}

 & \multirow[t]{9}{*}{Gemini-2.0-Flash} & P1 & {\textcolor{ForestGreen}{12.7***}} & {\textcolor{ForestGreen}{26.3***}} & {\textcolor{ForestGreen}{28.3***}} & 3.7 & {\textcolor{ForestGreen}{55.0***}} & {\textcolor{ForestGreen}{13.0***}} & {\textcolor{ForestGreen}{14.7***}} & {\textcolor{ForestGreen}{25.7***}} & {\textcolor{ForestGreen}{23.7***}} \\
 &  & P2 & - & {\textcolor{ForestGreen}{13.7**}} & {\textcolor{ForestGreen}{15.7**}} & {\textcolor{red}{-9.0*}} & {\textcolor{ForestGreen}{42.3***}} & 0.3 & 2.0 & {\textcolor{ForestGreen}{13.0***}} & {\textcolor{ForestGreen}{11.0**}} \\
 &  & P3 & - & - & 2.0 & {\textcolor{red}{-22.7***}} & {\textcolor{ForestGreen}{28.7***}} & {\textcolor{red}{-13.3**}} & {\textcolor{red}{-11.7**}} & -0.7 & -2.7 \\
 &  & P4 & - & - & - & {\textcolor{red}{-24.7***}} & {\textcolor{ForestGreen}{26.7***}} & {\textcolor{red}{-15.3**}} & {\textcolor{red}{-13.7**}} & -2.7 & -4.7 \\
 &  & P5 & - & - & - & - & {\textcolor{ForestGreen}{51.3***}} & {\textcolor{ForestGreen}{9.3**}} & {\textcolor{ForestGreen}{11.0*}} & {\textcolor{ForestGreen}{22.0***}} & {\textcolor{ForestGreen}{20.0***}} \\
 &  & P6 & - & - & - & - & - & {\textcolor{red}{-42.0***}} & {\textcolor{red}{-40.3***}} & {\textcolor{red}{-29.3***}} & {\textcolor{red}{-31.3***}} \\
 &  & P7 & - & - & - & - & - & - & 1.7 & {\textcolor{ForestGreen}{12.7***}} & {\textcolor{ForestGreen}{10.7*}} \\
 &  & P8 & - & - & - & - & - & - & - & {\textcolor{ForestGreen}{11.0**}} & {\textcolor{ForestGreen}{9.0*}} \\
 &  & P9 & - & - & - & - & - & - & - & - & -2.0 \\

\midrule \midrule

\multirow[t]{18}{*}{Dogs} & \multirow[t]{9}{*}{GPT-4o} & P1 & {\textcolor{red}{-10.0**}} & {\textcolor{ForestGreen}{10.0**}} & {\textcolor{ForestGreen}{13.7***}} & 5.0 & {\textcolor{ForestGreen}{12.3***}} & -6.0 & -5.3 & {\textcolor{ForestGreen}{8.0*}} & -3.3 \\
 &  & P2 & - & {\textcolor{ForestGreen}{20.0***}} & {\textcolor{ForestGreen}{23.7***}} & {\textcolor{ForestGreen}{15.0***}} & {\textcolor{ForestGreen}{22.3***}} & 4.0 & 4.7 & {\textcolor{ForestGreen}{18.0***}} & 6.7 \\
 &  & P3 & - & - & 3.7 & -5.0 & 2.3 & {\textcolor{red}{-16.0***}} & {\textcolor{red}{-15.3***}} & -2.0 & {\textcolor{red}{-13.3***}} \\
 &  & P4 & - & - & - & {\textcolor{red}{-8.7**}} & -1.3 & {\textcolor{red}{-19.7***}} & {\textcolor{red}{-19.0***}} & {\textcolor{red}{-5.7*}} & {\textcolor{red}{-17.0***}} \\
 &  & P5 & - & - & - & - & {\textcolor{ForestGreen}{7.3*}} & {\textcolor{red}{-11.0*}} & {\textcolor{red}{-10.3**}} & 3.0 & {\textcolor{red}{-8.3**}} \\
 &  & P6 & - & - & - & - & - & {\textcolor{red}{-18.3***}} & {\textcolor{red}{-17.7***}} & -4.3 & {\textcolor{red}{-15.7***}} \\
 &  & P7 & - & - & - & - & - & - & 0.7 & {\textcolor{ForestGreen}{14.0**}} & 2.7 \\
 &  & P8 & - & - & - & - & - & - & - & {\textcolor{ForestGreen}{13.3***}} & 2.0 \\
 &  & P9 & - & - & - & - & - & - & - & - & {\textcolor{red}{-11.3***}} \\

\noalign{\vskip 2pt} \cline{2-12} \noalign{\vskip 4pt}

 & \multirow[t]{9}{*}{Gemini-2.0-Flash} & P1 & 6.3 & {\textcolor{ForestGreen}{26.0***}} & 2.4 & 3.3 & {\textcolor{ForestGreen}{18.6***}} & 0.0 & {\textcolor{red}{-7.0**}} & {\textcolor{ForestGreen}{16.7***}} & {\textcolor{ForestGreen}{8.3*}} \\
 &  & P2 & - & {\textcolor{ForestGreen}{19.7***}} & -3.1 & -3.0 & {\textcolor{ForestGreen}{12.4**}} & -6.3 & {\textcolor{red}{-13.3**}} & {\textcolor{ForestGreen}{10.3*}} & 2.8 \\
 &  & P3 & - & - & {\textcolor{red}{-22.1***}} & {\textcolor{red}{-22.7***}} & {\textcolor{red}{-8.6*}} & {\textcolor{red}{-26.0***}} & {\textcolor{red}{-33.0***}} & {\textcolor{red}{-9.3*}} & {\textcolor{red}{-17.6***}} \\
 &  & P4 & - & - & - & 0.3 & {\textcolor{ForestGreen}{16.1***}} & -2.1 & {\textcolor{red}{-9.7**}} & {\textcolor{ForestGreen}{13.8***}} & 5.4 \\
 &  & P5 & - & - & - & - & {\textcolor{ForestGreen}{15.2***}} & -3.3 & {\textcolor{red}{-10.3***}} & {\textcolor{ForestGreen}{13.3***}} & 5.5 \\
 &  & P6 & - & - & - & - & - & {\textcolor{red}{-18.6***}} & {\textcolor{red}{-25.5***}} & -1.4 & {\textcolor{red}{-8.9**}} \\
 &  & P7 & - & - & - & - & - & - & {\textcolor{red}{-7.0*}} & {\textcolor{ForestGreen}{16.7***}} & 8.3 \\
 &  & P8 & - & - & - & - & - & - & - & {\textcolor{ForestGreen}{23.7***}} & {\textcolor{ForestGreen}{14.8***}} \\
 &  & P9 & - & - & - & - & - & - & - & - & {\textcolor{red}{-9.0*}} \\

\bottomrule
\end{tabular}
    
    \caption{Mean model performance difference between human pair 1 and pair 2 when doing the overhearer matching task based on the first rounds of conversations for 30 runs, each with a different object ordering. We use paired t-tests to determine if there is a significant difference between each human pair and highlight both \textbf{significant} \textcolor{ForestGreen}{positive} and \textcolor{red}{negative} mean differences. We indicate significance level using asterisks: ``*'' means \textit{p} < 0.05, ``**'' means \textit{p} < 0.01, and ``**'' means \textit{p} < 0.001. In each comparison, there are 30 experiments for each human pair, so the degree of freedom is 59.}
    \label{tab:meanDifferencesAcrossPairsOnR1}
\end{table*}

\begin{table*}[]
    \centering
    \small

\begin{tabular}{lllrrrr}
\toprule
 &  &  & Kendall Tau & Kendall \textit{p}-value & Spearman R & Spearman \textit{p}-value \\
Source & Model & Feature &  &  &  &  \\
\midrule
\multirow[t]{10}{*}{Baskets} 

 & \multirow[t]{5}{*}{Gemini-2.0-Flash} & \# Words & -0.02 & 1.00 & 0.02 & 0.96 \\

& & \# Sentences & -0.38 & 0.16 & -0.52 & 0.13 \\
 &  & \# Utterances & -0.16 & 0.60 & -0.18 & 0.63 \\
&  & \# Director Turns & -0.37 & 0.15 & -0.46 & 0.18 \\
 &  & \# Matcher Turns & -0.40 & 0.11 & -0.49 & 0.15 \\

\noalign{\vskip 2pt} \cline{2-7} \noalign{\vskip 4pt}

& \multirow[t]{5}{*}{GPT-4o}  &  \# Words & 0.11 & 0.73 & 0.19 & 0.60 \\
& & \# Sentences & -0.16 & 0.60 & -0.18 & 0.63 \\
 &  & \# Utterances & -0.02 & 1.00 & -0.04 & 0.91 \\
 &  & \# Director Turns & -0.18 & 0.47 & -0.24 & 0.51 \\
 &  & \# Matcher Turns & -0.22 & 0.37 & -0.29 & 0.41 \\

\midrule

\multirow[t]{10}{*}{Dogs} 

 & \multirow[t]{5}{*}{Gemini-2.0-Flash} & \# Words & 0.18 & 0.47 & 0.29 & 0.41 \\
 & & \# Sentences & -0.04 & 0.86 & -0.01 & 0.99 \\
 &  & \# Utterances & 0.18 & 0.47 & 0.24 & 0.50 \\
 &  & \# Director Turns & 0.18 & 0.47 & 0.25 & 0.49 \\
 &  & \# Matcher Turns & 0.18 & 0.47 & 0.24 & 0.50 \\

\noalign{\vskip 2pt} \cline{2-7} \noalign{\vskip 4pt}

 & \multirow[t]{5}{*}{GPT-4o} & \# Words & 0.38 & 0.16 & 0.53 & 0.12 \\
& & \# Sentences & 0.24 & 0.38 & 0.26 & 0.47 \\
 &  & \# Utterances & 0.38 & 0.16 & 0.50 & 0.14 \\
 &  & \# Director Turns & 0.20 & 0.48 & 0.28 & 0.43 \\
 &  & \# Matcher Turns & 0.16 & 0.53 & 0.21 & 0.57 \\

\bottomrule
\end{tabular}
    
    \caption{Correlations between average model performance based the 30 runs of an LVLM on R1 transcripts and the features of the related transcripts. None of these correlations are significant.}
    \label{tab:correlationAnalysesAvgPerformanceVersusInfoDensity}
\end{table*}

\subsection{Starting Early Versus Starting Late\label{app:earlyLateStartComp}}

Table~\ref{tab:meanDiffBetweenOverlappingRounds} shows the pairwise mean differences of overlapping rounds between an early start and a late start. Here, the overlap rounds can be one round or multiple rounds. For example, the overlapping rounds between R1-R4 and R2-R4 are R2-R4, whereas the verlapping round between R1-R4 and R4 is just R4. 

Table~\ref{tab:startingRoundsDifsTTestsBaskets} shows the pairwise mean differences between an early start and a late start. Here, we are comparing two single starting rounds, such as R1 versus R2, R1 versus R4 etc.

\subsection{Performance Difference between Human Pairs\label{app:meanDifferenceAcrossPairs}}

Table~\ref{tab:meanDifferencesAcrossPairsOnR1} shows the mean model performance difference between different human pairs when doing the overhearer matching task based on the first rounds of conversations. We use paired t-tests to determine a significant difference. 

We hypothesize that the performance variations may be due to different levels of information density in transcripts of different human pairs. We use five simple features to measure information density of a transcript: number of words, number of sentences, number of utterances, number of director turns, and number of matcher turns. We correlate these features with the average model performance and find that none of these correlations are statistically significant. See Table~\ref{tab:correlationAnalysesAvgPerformanceVersusInfoDensity} for details.

\subsection{Error Analyses\label{app:errorAnalyses}}

\begin{figure*}
    \centering
    \includegraphics[width=\linewidth]{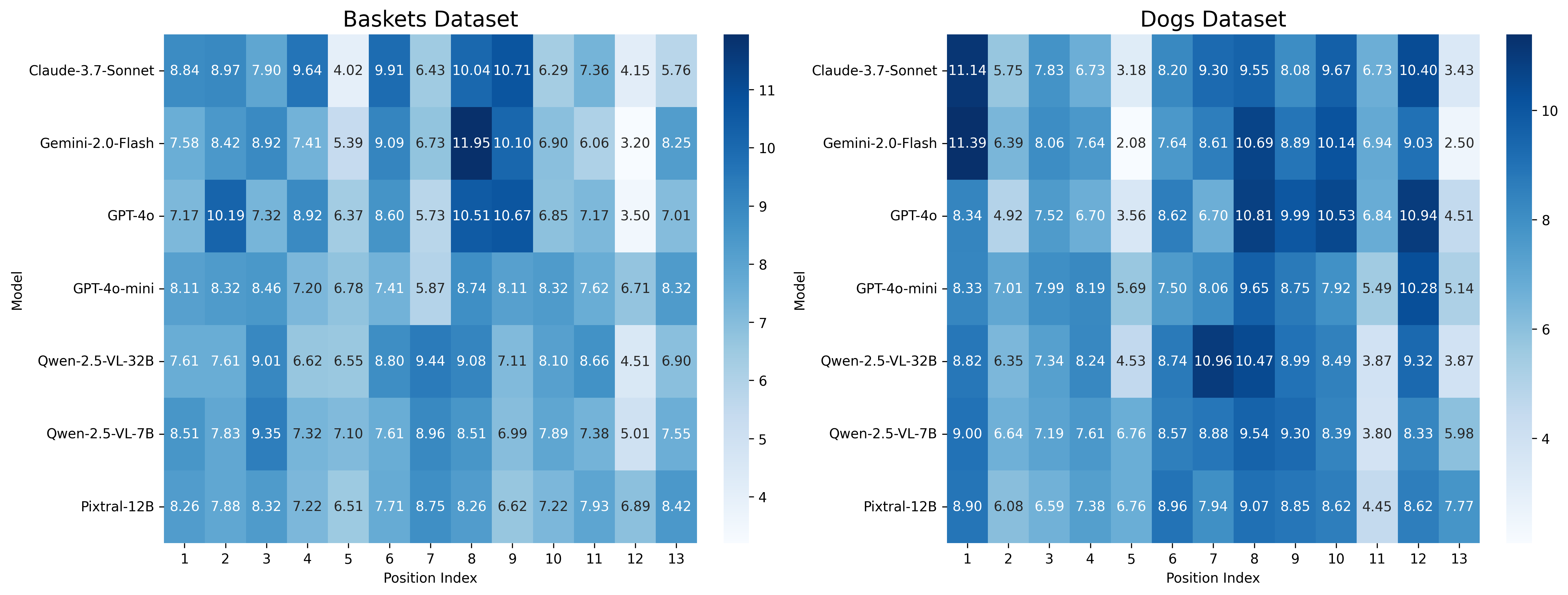}
    \caption{Percentage of times an LVLM fails to identify the correct object at index $i$ for the two datasets. The position index ranges from 1 to 13, since there are 13 objects for the overhearer and the target objects are shuffled across these positions.}
    \label{fig:positional_error_heatmap_r1_r4}
\end{figure*}

\begin{figure*}
    \centering
    \includegraphics[width=\linewidth]{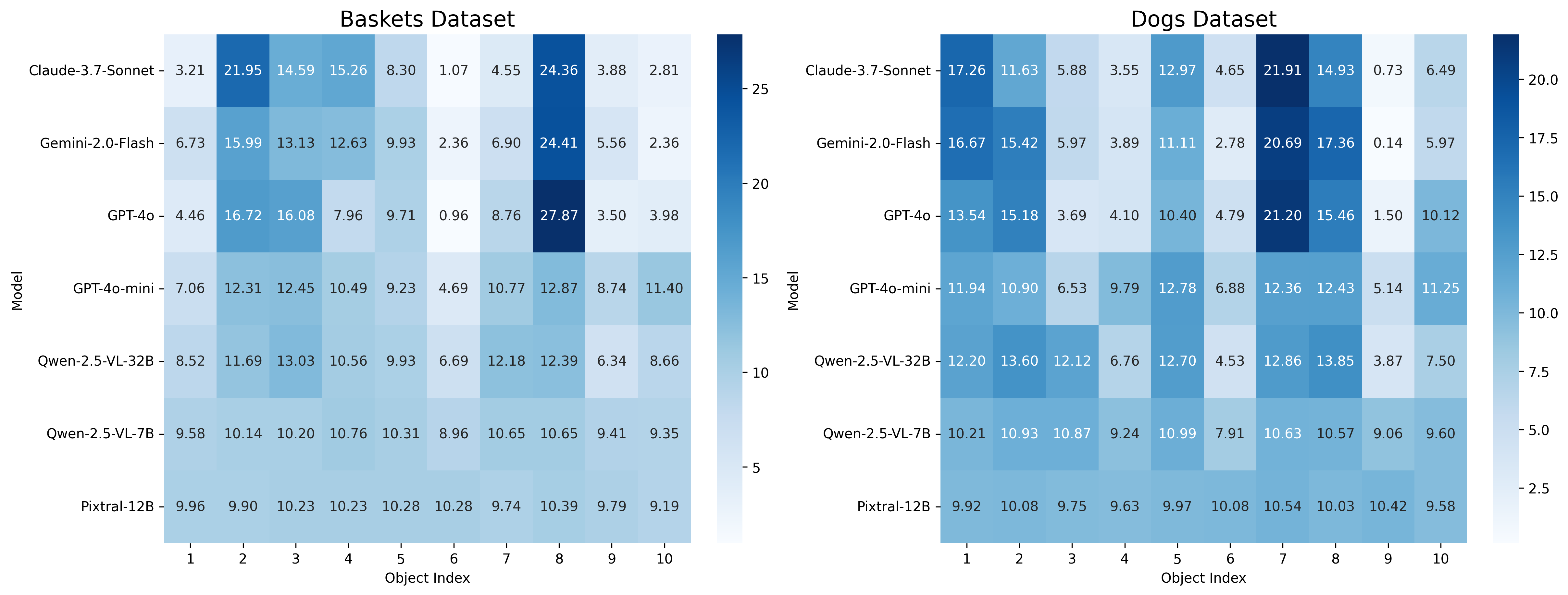}
    \caption{Percentage of times an LVLM fails to identify each one of the 10 target objects for the two datasets. The object index corresponds to those in Figure~\ref{fig:baskets} for baskets and in Figure~\ref{fig:dogs} for dogs.}
    \label{fig:object_error_heatmap_r1_r4}
\end{figure*}

Figure~\ref{fig:positional_error_heatmap_r1_r4} and  Figure~\ref{fig:object_error_heatmap_r1_r4} shows the percentage of time an LVLM fails to identify a target object placed at a position index $i$ across 13 positions and the specific object among 10 target objects for the two datasets, respectively. The two types of analyses are based only on experiments from R1 through R4 to control the effect of different starting rounds. However, our visualizations based on all the main experiments described in Section~\ref{sec:methodology} show similar patterns.

\clearpage

\definecolor{SeaGreen}{RGB}{46, 139, 87}      %
\definecolor{Periwinkle}{HTML}{CCCCFF}

\section{Prompt Templates\label{app:prompts}}

This section provides prompt templates for all experiments in this study. We use ``\$'' followed by a word to denote a placeholder.

\subsection{Prompt Template for Main Experiments}

Figure~\ref{fig:one_transcript_at_a_time_multi_turn} shows the prompt template used for the main experiments in Section~\ref{sec:results}. Each prompt contains a \hlc[SeaGreen!10]{system prompt} and a \hlc[Periwinkle!30]{user prompt}, where the \hlc[SeaGreen!10]{system prompt} is provided only once in the very beginning, whereas the \hlc[Periwinkle!30]{user prompt} may be used multiple times in case of a multi-turn conversation (the starting round is other than R4). Note that in the \hlc[SeaGreen!10]{system prompts}, ``\$example\_sequence'' (default=``3, 7, 1, 12, 5, 2, 13, 8, 10, 4, 6, 9, 11'') is fixed for a given value of ``\$num\_of\_objects'' (default=13) to maximize prompt similarity across different prompting conditions.

\definecolor{Lavender}{HTML}{E6E6FA}
\definecolor{Periwinkle}{HTML}{CCCCFF}
\definecolor{SeaGreen}{RGB}{46, 139, 87}
\definecolor{SkyBlue}{RGB}{135, 206, 235}
\definecolor{Orange}{HTML}{fdae61}
\definecolor{Brown}{HTML}{dfc27d}

\begin{figure*}[!ht]
\centering
% \small

\begin{tcolorbox}[
    width=\textwidth,
    colback=white,
    colframe=black,
    arc=4mm,
    boxrule=0.5pt,
    left=2mm,
    right=2mm,
    top=2mm,
    bottom=2mm,
    fonttitle=\bfseries,
    ]
    
\begin{tcolorbox}[
    colback=SeaGreen!8,
    boxrule=0pt,
    colframe=white,
    left=0pt,
    right=0pt,
    top=0pt,
    bottom=0pt,
    ]
% \small

You are an overhearer of a conversation between two participants engaged in a collaborative object-matching task for one or multiple rounds. Each participant is in a separate room and has a duplicate set of pictures arranged in different random orders. They cannot see each other’s sets and communicate solely via an audio link. During the task, one participant acts as the Director (D) and the other as the Matcher (M). The Director describes the pictures one at a time, and the Matcher selects the corresponding picture from their own set. Please note that it is the same two participants playing the same roles for all the rounds if there are multiple rounds. \newline

As the overhearer and for each round, you are provided with: \newline

- The full transcript of their conversation for that round.

- An image showing all pictures used in the task, randomly arranged and labeled with indices from 1 to \$num\_of\_objects. The image for each round may be different. \newline

Your goal is to determine the correct sequence of picture indices as described by the Director during each round. To do this: \newline

1. Carefully analyze the transcript to understand which pictures the Director refers to, in the order they were described.

2. Use the image to match each described picture to its corresponding index.

3. Think step by step and revise your reasoning and answers as needed. However, you may not ask questions or make assumptions beyond the given materials. \newline

When you reach your conclusion, output your response in the following format: \newline

Final Answer: [\$num\_of\_objects picture indices in correct order, separated by commas] \newline

Example: \$example\_sequence

\end{tcolorbox}
\begin{tcolorbox}[
    colback=Periwinkle!20,
    boxrule=0pt,
    colframe=white,
    left=0pt,
    right=0pt,
    top=0pt,
    bottom=0pt,
    ]
% \small

The transcript of the current conversation is as follows: \newline

\$transcript \newline

The image for the current round showing the pictures is as follows: \newline

<\$image\_path>

\end{tcolorbox}

\end{tcolorbox}
\caption{Prompt template for the experiments where the transcripts are presented one at a time in a multi-turn conversation. The \hlc[Periwinkle!30]{user prompt} at the bottom is repeated up to 4 times before each transcript/image pair in our experiments.}
\label{fig:one_transcript_at_a_time_multi_turn}

\end{figure*}

\definecolor{Lavender}{HTML}{E6E6FA}
\definecolor{Periwinkle}{HTML}{CCCCFF}
\definecolor{SeaGreen}{RGB}{46, 139, 87}
\definecolor{SkyBlue}{RGB}{135, 206, 235}
\definecolor{Orange}{HTML}{fdae61}
\definecolor{Brown}{HTML}{dfc27d}

\begin{figure*}[!ht]
\centering
% \small

\begin{tcolorbox}[
    width=\textwidth,
    colback=white,
    colframe=black,
    arc=4mm,
    boxrule=0.5pt,
    left=2mm,
    right=2mm,
    top=2mm,
    bottom=2mm,
    fonttitle=\bfseries,
    ]
    
\begin{tcolorbox}[
    colback=SeaGreen!8,
    boxrule=0pt,
    colframe=white,
    left=0pt,
    right=0pt,
    top=0pt,
    bottom=0pt,
    ]
% \small

You are an overhearer of a conversation between two participants engaged in a collaborative object-matching task for one or multiple rounds. Each participant is in a separate room and has a duplicate set of pictures arranged in different random orders. They cannot see each other"s sets and communicate solely via an audio link. During the task, one participant acts as the Director (D) and the other as the Matcher (M). The Director describes the pictures one at a time, and the Matcher selects the corresponding picture from their own set. Please note that it is the same two participants playing the same roles for all the rounds if there are multiple rounds. \newline

As the overhearer and for each round, you are provided with: \newline

- 10 object summaries based on the Director's description of the target pictures for that round. 

- An image showing all pictures used in the task, randomly arranged and labeled with indices from 1 to \$num\_of\_objects. The image for each round may be different. \newline

Your goal is to determine the correct sequence of picture indices as described by the Director during each round. To do this: \newline

1. Carefully analyze the transcript to understand which pictures the Director refers to, in the order they were described.

2. Use the image to match each described picture to its corresponding index.

3. Think step by step and revise your reasoning and answers as needed. However, you may not ask questions or make assumptions beyond the given materials. \newline

When you reach your conclusion, output your response in the following format: \newline

Final Answer: [\$num\_of\_objects picture indices in correct order, separated by commas] \newline

Example: \$example\_sequence

\end{tcolorbox}
\begin{tcolorbox}[
    colback=Periwinkle!20,
    boxrule=0pt,
    colframe=white,
    left=0pt,
    right=0pt,
    top=0pt,
    bottom=0pt,
    ]
% \small

The 10 object summaries based on the Director's description are as follows: \newline

\$summaries \newline

The image for the current round showing the pictures is as follows: \newline

<\$image\_path>

\end{tcolorbox}

\end{tcolorbox}
\caption{Prompt template for the experiments where the summaries of 10 target object summaries from a given transcript are presented, instead of the original transcript (\textbf{-Interaction}), one at a time in a multi-turn conversation. The \hlc[Periwinkle!30]{user prompt} at the bottom is repeated up to 4 times before each transcript/image pair in our experiments.}
\label{fig:ten_object_summaries_at_a_time_multi_turn}

\end{figure*}

\definecolor{Lavender}{HTML}{E6E6FA}
\definecolor{Periwinkle}{HTML}{CCCCFF}
\definecolor{SeaGreen}{RGB}{46, 139, 87}
\definecolor{SkyBlue}{RGB}{135, 206, 235}
\definecolor{Orange}{HTML}{fdae61}
\definecolor{Brown}{HTML}{dfc27d}

\begin{figure*}[!ht]
\centering
% \small

\begin{tcolorbox}[
    width=\textwidth,
    colback=white,
    colframe=black,
    arc=4mm,
    boxrule=0.5pt,
    left=2mm,
    right=2mm,
    top=2mm,
    bottom=2mm,
    fonttitle=\bfseries,
    ]
    
\begin{tcolorbox}[
    colback=SeaGreen!8,
    boxrule=0pt,
    colframe=white,
    left=0pt,
    right=0pt,
    top=0pt,
    bottom=0pt,
    ]
% \small

You are an overhearer of an ongoing conversation between two participants engaged in a collaborative object-matching task. Each participant is in a separate room and has a duplicate set of pictures arranged in different random orders. They cannot see each other"s sets and they can communicate solely via an audio link. During the task, one participant acts as the Director (D) and the other as the Matcher (M). The Director describes the pictures one at a time, and the Matcher selects the corresponding picture from their own set. \newline

As the overhearer and for each target picture, you are provided with: \newline

- An image showing all pictures used in the task, randomly arranged and labeled with indices from 1 to \$num\_of\_objects.

- Conversation between the Director (D) and Matcher (M) where the Matcher indicates that they have selected a target picture.  \newline

Your goal is to determine the correct sequence of picture indices as described by the Director during the task. To do this: \newline

1. Carefully analyze each conversation to understand which pictures the Director refers to, in the order they were described.

2. Use the image to match each described picture to its corresponding index. \newline

3. Think step by step and revise your reasoning and answers as needed. However, you may not ask questions or make assumptions beyond the given materials. \newline

You should produce a target picture index for each conversation presented to you as your current best guess. Once all the 10 pictures have been selected by the Matcher, you should reach a final conclusion and output your response in the following format: \newline

Final Answer: [\$num\_of\_objects picture indices in correct order, separated by commas] \newline 

Example: \$example\_sequence

\end{tcolorbox}
\begin{tcolorbox}[
    colback=Periwinkle!20,
    boxrule=0pt,
    colframe=white,
    left=0pt,
    right=0pt,
    top=0pt,
    bottom=0pt,
    ]
% \small

The image showing the pictures is as follows:
\newline

<\$image\_path> \newline

The conversation between the Director (D) and Matcher (M) for the first target picture is as follows: \newline

\$conversation

\end{tcolorbox}

\end{tcolorbox}
\caption{Prompt template for the experiments where a complete and manually segmented description of a target object (\textbf{ObjectDesc}) is provided one at a time in a multi-turn conversation. The \hlc[Periwinkle!30]{user prompt} at the bottom is repeated up to 10 times before each one of the 10 target objects' descriptions extracted from the transcript.}
\label{fig:one_object_description_at_a_time}

\end{figure*}

\definecolor{Lavender}{HTML}{E6E6FA}
\definecolor{Periwinkle}{HTML}{CCCCFF}
\definecolor{SeaGreen}{RGB}{46, 139, 87}
\definecolor{SkyBlue}{RGB}{135, 206, 235}
\definecolor{Orange}{HTML}{fdae61}
\definecolor{Brown}{HTML}{dfc27d}

\begin{figure*}[!ht]
\centering
% \small

\begin{tcolorbox}[
    width=\textwidth,
    colback=white,
    colframe=black,
    arc=4mm,
    boxrule=0.5pt,
    left=2mm,
    right=2mm,
    top=2mm,
    bottom=2mm,
    fonttitle=\bfseries,
    ]
    
\begin{tcolorbox}[
    colback=SeaGreen!8,
    boxrule=0pt,
    colframe=white,
    left=0pt,
    right=0pt,
    top=0pt,
    bottom=0pt,
    ]
% \small

You are an overhearer of a conversation between two participants engaged in a collaborative object-matching task for multiple rounds. Each participant is in a separate room and has a duplicate set of pictures arranged in different random orders. They cannot see each other’s sets and communicate solely via an audio link. During the task, one participant acts as the Director (D) and the other as the Matcher (M). The Director describes the pictures one at a time, and the Matcher selects the corresponding picture from their own set. Please note that it is the same two participants playing the same roles for all the rounds. \newline

As the overhearer, you are provided with: \newline

- The full transcripts of their conversation for each round.

- Images showing all pictures used in the task for each round, randomly arranged and labeled with indices from 1 to \$num\_of\_objects. \newline

Your goal is to determine the correct sequence of picture indices as described by the Director for each round. To do this: \newline

1. Carefully analyze the transcript to understand which pictures the Director refers to, in the order they were described.

2. Use the image to match each described picture to its corresponding index.

3. Think step by step and revise your reasoning and answers as needed. However, you may not ask questions or make assumptions beyond the given materials. \newline

When you reach your conclusion, output your response in the following JSON format for each round: \newline

Final Answer: \{``Round i'': [\$num\_of\_objects picture indices in correct order, separated by commas]\} \newline

Example: \{``Round 1'': [\$example\_sequence], ...\}

\end{tcolorbox}
\begin{tcolorbox}[
    colback=Periwinkle!20,
    boxrule=0pt,
    colframe=white,
    left=0pt,
    right=0pt,
    top=0pt,
    bottom=0pt,
    ]
% \small

The transcript of the conversation during round\#\$ix is as follows: \newline

\$transcript \newline

The image  for the round\#\$ix showing the pictures is as follows: \newline

<\$image\_path>

\end{tcolorbox}

\end{tcolorbox}
\caption{Prompt template for the experiments where multiple transcripts are presented together (\textbf{AllTranscript}) in a single-turn conversation. While we repeat the \hlc[Periwinkle!30]{user prompt} at the bottom four times for the four rounds, they are concentrated together and passed to an LVLM all at once.}
\label{fig:all_transcripts_at_once}

\end{figure*}

\subsection{Prompt Templates for Follow-Up Experiments}

We prompted \textsc{gpt-4.1-2025-04-14} to remove the colloquial and interactive features from our corpus.

\subsubsection{Prompt Template for Removing Colloquial Features\label{app:plus-formal}}

\begin{quote}
\small

You are given an excerpt from a transcribed, spontaneous conversation between two individuals. Your task is to revise the excerpt to produce a clear, polished version of the dialogue that reads like formal written text. Transform any standalone words or phrases into complete, grammatically correct sentences where appropriate. Do not add any additional information or context or change the meaning of the text. Do not output anything other than the revised excerpt. \newline

Here is the excerpt:
\${excerpt}

Revised Excerpt:
    
\end{quote}

\subsubsection{Prompt Template for Creating Object Summaries\label{app:minus-interaction}}

\begin{quote}
\small

You are given an excerpt from a transcribed, spontaneous conversation between two individuals. Your task is to extract and concisely summarize all descriptions used to characterize a specific object mentioned in the excerpt. You must follow the instructions below: \newline

1. Preserve all relevant descriptive details.

2. Do not alter the meaning, add context, or introduce new information.

3. Your response must only include the final summary—do not include the original excerpt or any explanatory text. \newline

Excerpt: \newline

\${excerpt}

Summary of Object Descriptions:
    
\end{quote}

\subsubsection{Prompt Template for Providing Transcripts without Colloquial Features}

We re-use the same prompt template from Figure~\ref{fig:one_transcript_at_a_time_multi_turn} when providing LVLMs with transcripts with colloquial features removed (\textbf{+Formal}). 

Appendix~\ref{app:plus-formal} shows the prompt template used for removing colloquial features from a transcript.

\subsubsection{Prompt Template for Providing Object Summaries}

Figure~\ref{fig:ten_object_summaries_at_a_time_multi_turn} shows the prompt template we used for providing LVLMs with all 10 target object summaries at once, instead of the original transcript \textbf{-Interaction}). Each prompt contains a \hlc[SeaGreen!10]{system prompt} and a \hlc[Periwinkle!30]{user prompt}, where we repeat the \hlc[Periwinkle!30]{user prompt} four times, containing 10 object summaries and the related input image from each round in order. We then concatenate these repeated user prompts together to prompt the LVLMs.

Appendix~\ref{app:minus-interaction} shows the prompt template used for creating object summaries from a given transcript.

\subsubsection{Prompt Template for Providing One Object Description at a time}

Figure~\ref{fig:one_object_description_at_a_time} shows the prompt template we used for providing a complete and manually segmented description of a target object (\textbf{ObjectDesc}) one at a time.

\subsubsection{Prompt Template for Providing All Transcripts At Once}

Figure~\ref{fig:all_transcripts_at_once} shows the prompt template we used for providing LVLMs with all transcripts at once (\textbf{AllTranscripts}). Each prompt contains a \hlc[SeaGreen!10]{system prompt} and a \hlc[Periwinkle!30]{user prompt}, where we repeat the \hlc[Periwinkle!30]{user prompt} four times, containing the transcript and the related input image from each round in order. We then concatenate these repeated user prompts together to prompt the LVLMs.

\subsubsection{Prompt Template for Providing LVLMs with Feedback}

We re-use the same prompt we used in the main experiments in Section~\ref{sec:results}, as shown in Figure~\ref{fig:one_transcript_at_a_time_multi_turn}. After an LVLMs produces its answer to each round, we insert the following prompt with the correct target sequence before proceeding to the next round of the matching task.

\begin{quote}
\small

Here is correct sequence of picture indices as described by the Director: \$answer. Reflect on your previous answer if it was wrong. We will proceed after your reflection.
\end{quote}

\end{document}